\pdfoutput=1

\documentclass[11pt]{article}

\usepackage[]{acl}

\usepackage{times}
\usepackage{latexsym}

\usepackage[T1]{fontenc}
\usepackage{color}
\usepackage{xcolor,colortbl}
\definecolor{green}{rgb}{0.1,0.1,0.1}
\definecolor{aliceblue}{rgb}{0.94, 0.97, 1.0}
\definecolor{Gray}{gray}{0.9}
\definecolor{GrAy}{gray}{0.96}
\definecolor{beaublue}{rgb}{0.74, 0.83, 0.9}
\definecolor{LightCyan}{rgb}{0.88,1,1}
\definecolor{inchworm}{rgb}{0.7, 0.93, 0.36}
\usepackage{comment}
\usepackage{xspace}
\usepackage{mathtools}
\usepackage{fontawesome}
\usepackage{adjustbox}
\usepackage{booktabs}
\usepackage{graphicx}
\usepackage{enumitem}
\usepackage{color}
\usepackage{latexsym}
\usepackage{graphicx}
\usepackage{amsmath}
\usepackage{mathtools}
\usepackage{breqn}
\usepackage{amsmath}
\usepackage{url}
\usepackage{enumerate}
\usepackage{amsmath}
\usepackage{amssymb}
\usepackage{comment}
\usepackage{xspace}
\usepackage{adjustbox}
\usepackage{booktabs}
\usepackage{graphicx}
\usepackage{enumitem}
\usepackage{color}
\usepackage[normalem]{ulem}
\usepackage{setspace}
\usepackage{relsize}

\usepackage{booktabs}
\usepackage{nicematrix}
\usepackage{xcolor}

\newcommand{\DrawPercentageM}[1]{%
  \begin{tikzpicture}
    \fill[color=blue!80,opacity=0.3]   (0.0 , 0.0) rectangle (#1*3ex , 1.5ex );
    \fill[color=gray!30] (#1*3ex  , 0.0) rectangle (3.0ex, 1.5ex);
  \end{tikzpicture}%
}

\newcommand{\DrawPercentageBarW}[1]{%
  \begin{tikzpicture}
    \fill[color=red!40, fill opacity=0.3]   (0.0 , 0.0) rectangle (#1*3ex , 1.5ex );
    \fill[color=gray!30] (#1*3ex  , 0.0) rectangle (3.0ex, 1.5ex);
  \end{tikzpicture}%
}

\newcolumntype{C}{ >{\vspace{-0.3cm}\centering\arraybackslash} m{4cm} }
\newcolumntype{D}{ >{\vspace{-0.1000cm}\centering\arraybackslash} m{1cm} }

 \newcolumntype{W}[1]{>{\centering\let\newline\\\arraybackslash\hspace{0pt}}m{#1}}
\newcolumntype{L}[1]{>{\raggedright\let\newline\\\arraybackslash\hspace{0pt}}m{#1}}
\usepackage{amssymb}
\usepackage{pifont}
\newcommand{\xmark}{\ding{55}}%

\usepackage{xcoffins}
\NewCoffin\tablecoffin
\NewDocumentCommand\Vcentre{m}
  {%
    \SetHorizontalCoffin\tablecoffin{#1}%
    \TypesetCoffin\tablecoffin[l,vc]%
  }
  \newcolumntype{C}[1]{>{\centering\let\newline\\\arraybackslash\hspace{0pt}}m{#1}}
\newcolumntype{L}[1]{>{\raggedright\let\newline\\\arraybackslash\hspace{0pt}}m{#1}}

\def\eg{\emph{e.g}\onedot} 
\def\ie{\emph{i.e}\onedot} 
 
\def\etc{\emph{etc}\onedot}

\makeatother

\DeclareRobustCommand\ie{%
  \UKUS@comma{i.e}%
}
\DeclareRobustCommand\eg{%
  \UKUS@comma{e.g}%
}
\definecolor{lightcyan}{rgb}{0.88, 1.0, 1.0}
 	\definecolor{cyan}{rgb}{0.0, 1.0, 1.0}
\usepackage{tikz}

\newcommand{\DrawPercentageBar}[1]{%
  \begin{tikzpicture}
    \fill[color=black!40]   (0.0 , 0.0) rectangle (#1*3ex , 1.5ex );
    \fill[color=gray!30] (#1*3ex  , 0.0) rectangle (3.0ex, 1.5ex);
  \end{tikzpicture}%
}

\DeclareRobustCommand\onedot{\futurelet\@let@token\@onedot}
\def\@onedot{\ifx\@let@token.\else.\null\fi\xspace}
\newcommand{\specialcell}[2][c]{%
  \begin{tabular}[#1]{@{}c@{}}#2\end{tabular}}

\usepackage{xspace}

\newcolumntype{C}{ >{\vspace{-0.3cm}\centering\arraybackslash} m{4cm} }
\newcolumntype{D}{ >{\vspace{-0.1000cm}\centering\arraybackslash} m{1cm} }

\newcolumntype{C}{ >{\vspace{-0.3cm}\centering\arraybackslash} m{4cm} }
\newcolumntype{D}{ >{\vspace{-0.1000cm}\centering\arraybackslash} m{1cm} }

\newcommand{\DrawPercentageBarNeg}[1]{%
  \begin{tikzpicture}
    \fill[color=red!30]   (0.0 , 0.0) rectangle (#1*3ex , 1.5ex );
    \fill[color=gray!30] (#1*3ex  , 0.0) rectangle (3.0ex, 1.5ex);
  \end{tikzpicture}%
}
\makeatletter
\DeclareRobustCommand\onedot{\futurelet\@let@token\@onedot}
\def\@onedot{\ifx\@let@token.\else.\null\fi\xspace}

\def\eg{\emph{e.g}\onedot} 
\def\ie{\emph{i.e}\onedot} 
 
\def\etc{\emph{etc}\onedot} 
 
\def\simprob{\texttt{SimProb}}

\usepackage[utf8]{inputenc}

\usepackage{microtype}

\title{Belief Revision based Caption Re-ranker  \\ with Visual Semantic Information}

\author{Ahmed Sabir$^{1}$, Francesc Moreno-Noguer$^{2}$, Pranava Madhyastha$^{3}$, Llu\'{\i}s Padr\'o$^{1}$ \\
$^{1}$ Computer Science Department, Universitat Polit\`ecnica de Catalunya, Barcelona, Spain  \\ 
$^{2}$ Institut de Rob\`otica i Inform\`atica Industrial, CSIC-UPC, Barcelona, Spain\\
$^{3}$ City, University of London, London, UK 
}

\begin{document}
\maketitle
\begin{abstract}
In this work, we focus on improving the captions generated by image-caption generation systems. We propose a novel re-ranking approach that leverages visual-semantic measures to identify the ideal caption that maximally captures the visual information in the image. Our re-ranker utilizes the Belief Revision framework \cite{blok2003probability} to calibrate the original likelihood of the top-n captions by explicitly exploiting the semantic relatedness between the depicted caption and the visual context. Our experiments demonstrate the utility of our approach, where we observe that our re-ranker can enhance the performance of a typical image-captioning system without the necessity of any additional training or fine-tuning.\footnote{\url{https://github.com/ahmedssabir/Belief-Revision-Score}}
\end{abstract}

\section{Introduction}


Image caption generation is a task that predominantly lies at the intersection of the areas of computer vision and natural language processing. The task is primarily aimed at generating a natural language description for a given image. Caption generation systems usually consist of an image encoder that encodes a given image (usually by using a CNN) whose encoding is fed to a decoder (usually by using a generative model such as RNN) to generate a natural language sentence which describes the image succinctly. The most widely used approaches include a CNN-RNN end-to-end system \cite{Oriol:15,anderson2018bottom}, end-to-end systems with attention that attend to specific regions of the image for generation \cite{xu2015show,you2016image} and systems with reinforcement learning based methods \cite{rennie2017self,ren2017deep}. Furthermore, recent advances have resulted in end-to-end systems that use Transformer based architecture for language generation and have become the current state-of-the-art \cite{huang2019attention,Marcella:20,zhang2021rstnet}.

\begin{figure}
\small 
\centering 

\includegraphics[width=0.99\columnwidth]{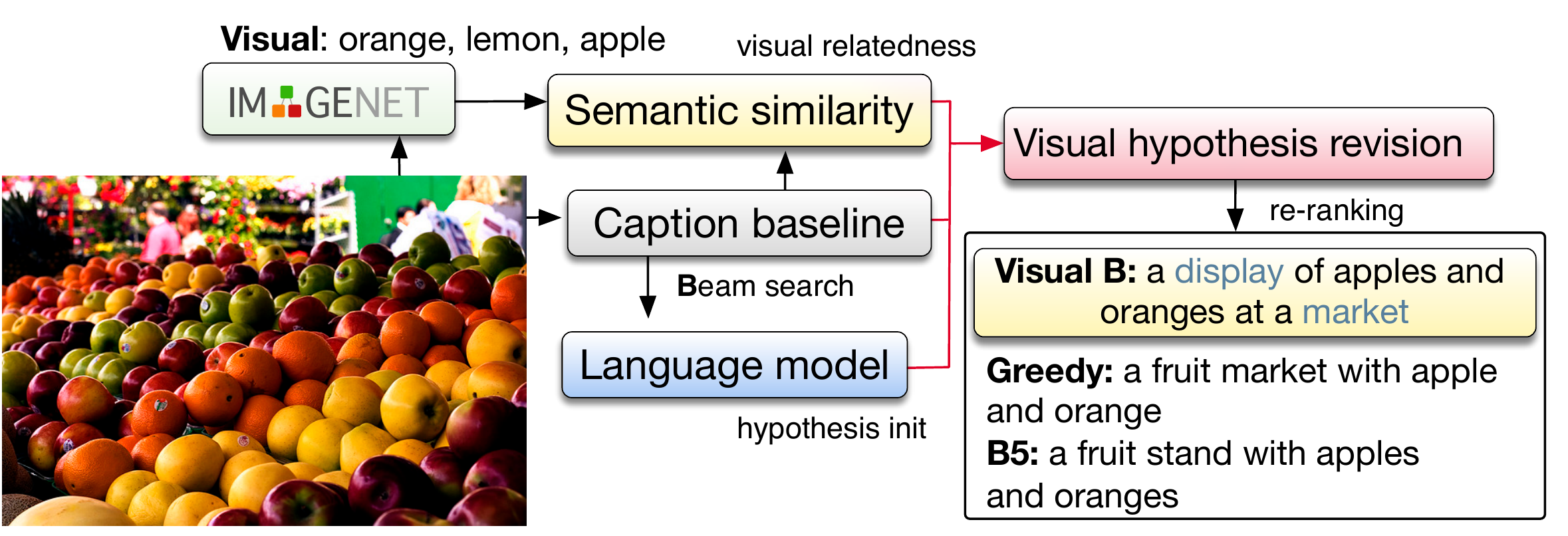} 

\vspace{-0.2cm}
\caption{An overview of our hypothesis revision based visual re-ranker. We use the visual context from the image to \textit{revise} and re-rank  the most closely related caption to its visual context. These semantic relatedness measures are learned at the word-to-sentence level. In this example, we showcase our visual re-ranker (\textbf{V}isual \textbf{B}eam), a post-processing approach, which is able to re-rank the most `descriptive caption' from the $5$-Best \textbf{B}eam \cite{Marcella:20}.}

\label{fig:main figure}
 \end{figure}


While the state-of-the-art models generate captions that are comparable to human level captions, they are known to lack lexical diversity, are often not very distinct, and sound synthetic. We here highlight a few recent approaches that have focused on this problem, these include \citet{dai2017towards} that uses generative adversarial networks towards generating diverse and human like captions. \citet{vedantam2017context} use a beam search with a distractor image to force the model to produce diverse captions by encouraging the models to be discriminative. Other recent works use a beam search directly to produce diverse captions by forcing richer lexical word choices \cite{ippolito2019comparison,vijayakumar2018diverse,wang2019describing,wang2020compare}. In this work, we follow a similar line of research and focus on the problem of improving diversity and making captions natural and human like and propose a novel re-ranking approach. In this approach, we use $n$-best reranking with a given beam that explicitly uses the semantic correlation between the caption and the visual context through belief revision (an approach inspired by human logic). We refer the reader to Figure \ref{fig:main figure}, where the approach results in a caption that is a) visually relevant and b) the most natural and human like.

Our primary contributions in this paper are:

\begin{itemize}[noitemsep]
    \item We demonstrate the utility of the Belief Revision \cite{blok2003probability} framework, which has been shown to correlate highly  with human judgment and has demonstrate its applicability to the task of Image Captioning. We do this by employing vision-language joint semantic measures using state-of-the-art pre-trained language models. 

    \item Our approach is a \emph{post-processing} method and is devised to be a drop-in replacement for any caption system.
    
    
    
    \item Through our experiments, we report that our proposal selects better captions as reported using automated metrics, as well as being validated by human evaluations.
    
    
\end{itemize}

\section{Belief Revision with \simprob{} Model}
\label{se:belife-revision}
In this section, we briefly introduce \simprob{}, which is based on the philosophical intuitions of Belief Revision, an idea that helps to convert similarity measures to probability estimates. \citet{blok2003probability} introduce a conditional probability model that assumes that the preliminary probability result is updated or revised to the degree that the hypothesis proof warrants. The range of revision is based on the informativeness of the argument and its degree of similarity. That is, the similarity to probability conversion can be defined in terms of \textbf{Belief Revision}. Belief Revision is a process of forming a belief by taking into account a new piece of information.


Let us consider the following statements: 
\begin{itemize}[noitemsep] 
\small 
\item[\texttt{1}] \texttt{Tigers can bite through wire, therefore  Jaguars  can bite through wire.}
\item[\texttt{2}] \texttt{Kittens can bite through wire, therefore  Jaguars can bite through wire.}

\end{itemize}

In the first case, the statement seems logical because it matches our prior belief \ie jaguars are similar to tigers, so we expect them to be able to do similar things. We hence consider that the statement is consistent with our previous belief, and there is no need to revise it. In the second case, the statement
is surprising because our prior belief is that kittens are not as similar to jaguars,
and thus, not so strong. But if we assume the veracity of the statement, then we
need to revise and update our prior belief about the strength of kittens.

 This work formalizes belief as probabilities and revised belief as conditional probabilities and provides a framework to compute them based on the similarities of the involved objects. According to the authors, belief revision should be
proportional to the similarity of the involved objects (\ie in the example, the statement about kittens and jaguars would cause a stronger belief revision than \eg the same statement involving pigeons and jaguars because they are less similar). In our case, we use the same rationale and the same formulas to convert similarity (or relatedness) scores into probabilities suitable for reranking.

\noindent{\bf \simprob{} Model} To obtain the likelihood revisions based on similarity scores, we need three parameters: (1) \textbf{Hypothesis}: prior probabilities, (2) \textbf{Informativeness}: conclusion events and (3) \textbf{Similarities}: measuring the relatedness between involved categories.The goal is to predict a conditional probability of statements, given one or more other statements. In order to predict the conditional probability of the argument’s
conclusion, given its premise or hypothesis, we will need only the prior probabilities of the statements, as well as the similarities between the involved categories (\eg kittens and tigers). 

\noindent{\bf Formulation of \simprob{}} The conditional probability $\operatorname{P}(Q_{c} | Q_{a})$ is expressed in terms of the prior probability of the conclusion statement $\operatorname{P}(Q_{c})$, the prior probability of the premise statement  $\operatorname{P}(Q_{a})$, and the similarity between the conclusion and the premise categories
sim(a, c). 

\resizebox{0.45 \textwidth}{!} 
{
$
\mathrm{P}\left(Q_{\mathrm{c}} \mid Q_{a}\right)=\mathrm{P}\left(Q_{c}\right)^{\alpha}{ }_{\text {where }} \alpha=\left[\frac{1-\operatorname{sim}(a, c)}{1+\operatorname{sim}(a, c)}\right]^{1-\mathrm{P}\left(Q_{a}\right)}
$
} 


\vspace{1mm}

\noindent{\bf Belief Revision Elements} As we discussed above, there are two factors that determine the hypothesis probability revision: 1) the sufficient relatedness to the category: as $sim(a,c){\to}0$, $\alpha{\to}1$, and thus $\operatorname{P}(Q_{c} | Q_{a}) = \operatorname{P}(Q_{c})$, \ie no revision takes place, as there are no changes in the original belief. While as $sim(a,c){\to}1$, $\alpha{\to}0$, and the hypothesis probability $\operatorname{P}(Q_{c})$ is revised and is raised closer to 1;  2) the informativeness of the new information $1-\operatorname{P}(Q_{a})$: as $\operatorname{P}(Q_{a}){\to}1$ and in consequence is less informative, $\alpha{\to}1$, as there is no new information, and hence no revision is required. 


\section{Visual Re-ranking for Image Caption}
\subsection{Problem Formulation} 

The beam search is the dominant method for approximate decoding in structured prediction tasks such as machine translation, speech recognition and image captioning. A larger beam size allows the model to perform a better exploration in the search space compared to greedy decoding. The main idea of the beam search is to explore all possible captions in the search space  by  keeping a set of \textit{top candidates}. 


Our goal is to leverage the visual context information of the image to re-rank the candidate sequences obtained through the beam search, thereby moving the most visually relevant candidate up in the list, as well as moving wrong candidates down. For this  purpose,  we  experiment  with  different  re-rankers, based on the relatedness between the candidate caption and the  semantic  context  observed  in  the  image through the idea of Belief Revision.  

\paragraph{Caption Extraction}

We employ two recent Transformer based architectures for caption generation to extract the top candidate captions using different beam sizes ($B=1\ldots{20}$) \cite{vijayakumar2018diverse}. The first baseline is based on a multi-task model for discriminative Vision and Language BERT \cite{lu202012} that is fine-tuned on 12 downstream tasks. The second baseline is the vanilla Transformer \cite{Ashish:17} with the Meshed-Memory based caption generator \cite{Marcella:20} with pre-computed top-down visual features \cite{anderson2018bottom}.  

\subsection{Proposal}



One approach of using word-level semantic relations for scene text correction with the visual context of an image was introduced in \citet{sabir2018visual}, which allows for the establishment of learning semantic correlations between a visual context and a text fragment.  In our work, this semantic relatedness is between a visual context and a given candidate caption (\ie beam search), and uses Belief Revision (BR) via \simprob{} to re-visit and re-rank the original beam search based on the similarity to the \emph{image objects/labels $c$} (a proxy for image context). The BR in this scenario is a conditional probability which assumes that the caption preliminary probability (\textit{hypothesis}) $\operatorname{P}(w)$ is revised to the degree approved by the semantic similarity with visual context $\operatorname{sim}(w, c)$. The final output caption $w$ for a given visual context $c$ is written as:





\label{sec:belief}

 
 

 \begin{equation}
\operatorname{P}(w \mid c)=\operatorname{P}(w)^{\alpha}
\label{ch2: sim-to-prob-1}
\end{equation}
\noindent  where the main components of visual based hypothesis revision:

\noindent{\textbf{Hypothesis}:} $\operatorname{P}(w)$ 

\noindent{\textbf{Informativeness}:}    $1-\operatorname{P}(c)$ 

\noindent{\textbf{Similarities}:}  $\alpha=\left[\frac{1-\operatorname{sim}(w, c)}{1+\operatorname{sim}(w, c)}\right]^{1-\operatorname{P}(c)}$ 


\vspace{1mm}
\noindent where $\operatorname{P}(w)$ is the \textit{hypothesis} probability (beam search candidate caption) and  $\operatorname{P}(c)$ is the probability of the evidence that causes hypothesis probability revision (visual context from the image). We next discuss the details of each component in \simprob{} as visual based re-ranker. 





\noindent{\bf Hypothesis:} Prior probabilities of original belief. As this approach is inspired by humans, the hypothesis $\operatorname{P}(w)$ needs to be initialized by a common observation such as a Language Model (LM) trained on a general text corpus. Therefore, we employ a Generative Pre-trained Transformer (GPT-2) \cite{radford2019language} a LM to initialize the hypothesis probability. We set $\operatorname{P}(w)$ as the mean of LM token probability. 

\noindent{\bf Informativeness:} Inversely related to the probability of set $\operatorname{P}(c)$ information that causes hypothesis revision. We leverage ResNet \cite{Kaiming:16} and an Inception-ResNet v2 based Faster R-CNN object detector \cite{huang2017speed}\footnote{TensorFlow Object Detection API}  to extract textual visual context information from the image. We use the classifier probability confidence with a threshold to filter out non-existent objects  in the image. \textcolor{black}{For each image, we extract  visual information as follows: (1) top-1 concept (2) multi concept top-3 (label class or object category) visual information. For the single concept,  we employ a unigram LM, based on the 3M-token opensubtitles corpus \cite{lison2016opensubtitles2016}, to initialize the informativeness of the visual information. For multiple concepts, we take the mean probability of the three concepts. Note that we are initializing the single visual context with LM to maximize the visual context score while computing the informativeness.}


\noindent{\bf Similarities:} Hypothesis revision is more likely if there is a close relation between the hypothesis and the new information (candidate caption and visual context in our case). We rely on two of the most recent state-of-the-art pre-trained Transformer-based language models to compute the semantic similarity between the caption and its visual context information with contextual embedding. In particular, we utilize the visual as context for the sentence (\ie caption) to compute the cosine distance: 

\begin{itemize}
   \itemsep 0cm
\item \textbf{BERT \cite{Jacob:19}:} BERT achieves remarkable results on many sentence level tasks and especially in the textual semantic similarity task (STS-B) \cite{cer2017semeval}. Therefore, we fine-tuned BERT$_{\text{base}}$ on the training dataset, (textual information, 460k captions: 373k for training and 87k for validation) \ie visual context, caption, label [semantically related or not related]), with a binary classification cross-entropy loss function [0,1] where the target is the semantic similarity between the visual and the candidate caption, with batch size 16 for two epochs with a learning rate $2\mathrm{e}{-5}$. 





\item \textbf{RoBERTa \cite{liu2019roberta}:}  RoBERTa is an improved version of BERT, trained on a large amount of data, using dynamic masking strategies to prevent overfitting. It achieves a 2.4\% improvement over BERT$_{\text{Large}}$ in the STS task. Since RoBERTa$_{\text{Large}}$ is more robust, we use an off-the-shelf model tuned on STS-B task. 
In particular, we follow the traditional approach to compute the semantic similarity with a BERT based model with a mean pool, over the last hidden layer, to extract a meaningful vector to compute the cosine distance.

\end{itemize}

\section{Experiments}

\subsection{Dataset}
\label{data:dataset}


\noindent{\bf COCO-Caption \cite{Tsung-Yi:14}:} This dataset  contains around 120k images and each image is annotated with five different human-written captions. We use the split provided by \cite{karpathy2015deep}, where 5k images are used for testing, 5k for validation, and the rest for model training.

 \begin{table*}[ht]
\small 
\centering
\resizebox{0.7\linewidth}{!}{
\begin{NiceTabular}{lccccccc}
\toprule   
 Model & B-1 & B-4 & M & R & C & S &BERTscore\\ 
\midrule
\multicolumn{8}{l}{VilBERT \cite{lu202012}}  \\
Vil$_{\text{Greedy}}$  & 0.751  & 0.330 & 0.272 & 0.554 &  1.104 & 0.207 &0.9352 \\


\addlinespace[0.1cm]
Vil$_{\text{BeamS}}$  &  0.752   &  0.351  &  0.274 &  0.557   & 1.115 & 0.205 & 0.9363  \\


\addlinespace[0.1cm]
Vil+VR$_{\text{W-Object}}$  \cite{fang2015captions} & 0.756 &  0.348  &  0.274 & 0.559 & 1.123 &0.206 &  0.9365 \\ 
Vil+VR$_{\text{Object}}$  \cite{wang2018object}   & 0.756    &  0.348  &  0.274 &  0.559   & 1.120 & 0.206 & 0.9364 \\ 
Vil+VR$_{\text{Control}}$   \cite{cornia2019show}  & 0.753  &  0.345  &  0.274 &  0.557  & 1.116 & 0.206 & 0.9361 \\ 
\addlinespace[0.1cm]
Vil+VR$_{\text{BERT}}$  (only $sim$)  & 0.753  &  0.343    &  0.273  & 0.556  & 1.112 & 0.206 & 0.9361  \\ 
\hline
Vil+VR$_{\text{BERT}}$   & 0.752 &  0.351 &  0.274    &  0.557  & 1.115 & 0.205 & 0.9365 \\

Vil+VR$_{\text{BERT-Object}}$ & 0.752 &   0.352  &   \textbf{0.277} & 0.560 &   1.129  & \textbf{0.208}  & 0.9365 \\  


Vil+VR$_{\text{RoBERTa}}$   & 0.753  &  \textbf{0.353} &  0.276  &   0.559   & 1.128  & 0.207 & \textbf{0.9366}  \\


 
Vil+VR$_{\text{RoBERTa-Object}}$ & \textbf{0.758}   &  0.344 & 0.262  &  0.555 &  1.234 &  0.206 & 0.9365   \\

\hline 
Vil+VR$_{\text{BERT-Multi-class}}$   & 0.753 &   \textbf{0.353} &  0.276    &  0.559  &  1.131 & \textbf{0.208} & 0.9365 \\
Vil+VR$_{\text{BERT-Multi-object}}$   &  0.752 &  0.351 &  0.276    &  0.558  &  1.123 &  \textbf{0.208} & 0.9364 \\
\rowcolor{Gray}
Vil+VR$_{\text{RoBERTa-Multi-class}}$  & 0.751 & 0.351 &   \textbf{0.277}    &  \textbf{0.561}   & \textbf{1.137}   & \textbf{0.208} &   \textbf{0.9366}  \\ 

Vil+VR$_{\text{RoBERTa-Multi-object}}$  & 0.752 &  \textbf{0.353} &   \textbf{0.277}    &  0.559   & 1.131  & \textbf{0.208} &    \textbf{0.9366}   \\ 






\midrule 
\addlinespace[0.1cm]
\multicolumn{8}{l}{Transformer based caption generator \cite{Marcella:20}}  \\

Trans$_{\text{Greedy}}$   & 0.787  & 0.368 & 0.276 & 0.574 &   1.211 & 0.215 & 0.9376 \\ 
\addlinespace[0.1cm]
    Trans$_{\text{\text{BeamS}}}$  & 0.793   &   0.387  &0.281 &  0.582   &  1.247 & \textbf{0.220} & \textbf{0.9399}  \\

\addlinespace[0.1cm]
Trans+VR$_{\text{W-Object}}$  \cite{fang2015captions} & 0.786  &  0.378  &  0.277 &   0.579 & 1.228 & 0.216 &  0.9388  \\ 
Trans+VR$_{\text{Object}}$  \cite{wang2018object}  & 0.790    &  0.383 &   0.280 &0.580 &  1.237   & 0.219 &  0.9391   \\ 
Trans+VR$_{\text{Control}}$   \cite{cornia2019show}  & 0.791    &  \textbf{0.388}  &  0.281 &  \textbf{0.583}  & 1.248 & \textbf{0.220} & 0.9398  \\ 

\addlinespace[0.1cm]
Trans+VR$_{\text{BERT}}$ (only $sim$) & 0.789  & 0.380   &  0.279  & 0.579 & 1.234 & 0.219 & 0.9389  \\ 
\hline 
\rowcolor{Gray}
Trans+VR$_{\text{BERT}}$ & 0.793  &  \textbf{0.388}   &  \textbf{0.282}  & \textbf{0.583} &   \textbf{1.250}  &  \textbf{0.220} & \textbf{0.9399}   \\  

Trans+VR$_{\text{BERT-Object}}$ & 0.793 &   0.385  &   0.281 &  0.581&   1.242  & 0.219  & 0.9396 \\  

Trans+VR$_{\text{RoBERTa}}$  & 0.792 & 0.386 & 0.280  &  0.582 & 1.244 & 0.219 & 0.9395 \\

Trans+VR$_{\text{RoBERTa-Object}}$ & 0.792 & 0.386 & 0.281  &  0.582 & 1.242 & 0.219 & 0.9396  \\

\hline 
Trans+VR$_{\text{BERT-Multi-class}}$   & \textbf{0.794} &  0.385 & 0.281    &   0.582  &  1.248 &  \textbf{0.220} & 0.9395 \\
Trans+VR$_{\text{BERT-Multi-object}}$   & 0.792 &  0.385 &  0.281    &   0.582  &   1.244 &  \textbf{0.220} & 0.9395 \\
Trans+VR$_{\text{RoBERTa-Multi-class}}$  & 0.791   & 0.385 &   0.280    &   0.581   & 1.244    &  0.219 &    0.9395    \\ 
Trans+VR$_{\text{RoBERTa-Multi-object}}$  & 0.791 & 0.385&  0.281    &   0.582   & 1.243   &  0.219 &   0.9395    \\


\bottomrule

\end{NiceTabular}
}
\vspace{-0.2cm}
\caption{Performance of compared baselines on the Karpathy test split with/without \textbf{V}isual semantic \textbf{R}e-ranking. For each base system, we report performance using a greedy search and the best beam search. Re-ranking is applied to the top-20 results of each system using BERT or RoBERTa for caption-context similarity. The visual contexts are extracted using ResNet152 and Inception Resnet v2 based Faster R-CNN \textbf{object} detector. We also report results for Bert-based similarity without a hypothesis probability (rows marked \textit{only sim}).}
\label{se:Table}
\end{table*}

 

\noindent{\bf{Visual Context Enrichment:}} We enrich COCO-Caption with textual visual context information. To automate visual context generation and without the need for a human label, for the training dataset, we use only ResNet152, which has 1000 label classes, to extract the top-k three label class visual context information for each image in the caption dataset.  For testing, we rely only on the top-k visual information as a concept, and we also employ the Inception-ResNet v2 based Faster R-CNN object detector with 80 object classes. In particular, each single annotated caption has three visual context information.



{\bf{\noindent{Evaluation Metric}:}} We use the official COCO offline evaluation suite, producing several widely used caption quality metrics: \textbf{B}LEU \cite{papineni2002bleu} \textbf{M}ETEOR \cite{banerjee2005meteor}, \textbf{R}OUGE  \cite{lin2004rouge}, \textbf{C}IDEr \cite{vedantam2015cider},  \textbf{S}PICE \cite{anderson2016spice} and BERTscore \cite{bert-score}.

\subsection{Results and Discussion}

We use visual semantic information to re-rank candidate captions produced by out-of-the-box state-of-the-art caption generators. We extract the top-20 beam search candidate captions from two state-of-the-art models:  VilBERT  \cite{lu202012}, fine-tuned on a total of 12 different vision and language datasets such as caption image retrieval and visual question answering, and a specialized caption-based Transformer \cite{Marcella:20}.

Experiments applying different rerankers to the each base system are shown in Table~\ref{se:Table}. 
The tested rerankers are: (1) VR$_{\text{BERT}}$ using BERT similarity between the candidate caption and the visual context of the image, transforming it to a probability using Equation \ref{ch2: sim-to-prob-1}, and combines the result with the original candidate probability to obtain the reranked score. (2) VR$_{\text{RoBERTa}}$ carrying out the same procedure using similarity produced by RoBERTa. A simpler model is also tested --VR$_{\text{BERT}}$ (only $sim$) in Table~\ref{se:Table}--, which replaces Equation  \ref{ch2: sim-to-prob-1} with $\operatorname{P}(w \mid c)=\operatorname{sim}(w, c)^{\operatorname{P}(c)}$, that is, it does not rely on the original caption probability.

\begin{figure}[t!]
\centering 

\includegraphics[width=\columnwidth]{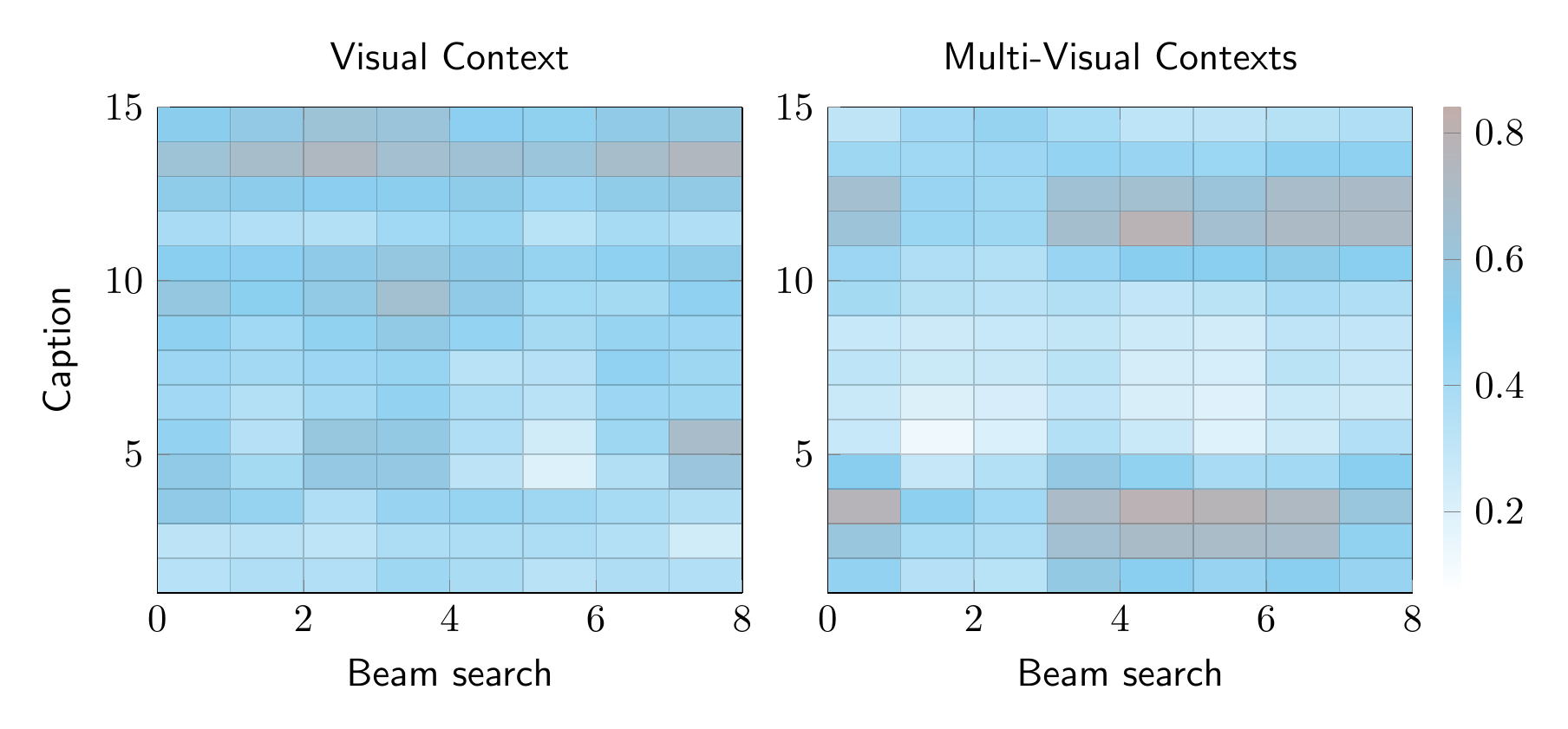}

\vspace{-0.3cm}

\caption{Visualization of top-$k$ nine re-ranked Beam search via \texttt{SimProb} with, Vil+VR$_{\text{RoBERTa}}$ (Right)  multiple visual and (Left) one concept visual context. The longer caption benefits from using multiple concepts. }  
\label{fig:muti}
\end{figure}

First, we compare our work with the original visual caption re-ranker with multiple word objects as concepts from the image, that are extracted via Inception-ResNet v2 based Faster RCNN (\ie person, van, \etc), $\text{VR}_{\text{w-Object}}$  \cite{fang2015captions}. However, to make a fair comparison, we use the Sentence-RoBERTa$_{\text{Large}}$ for the sentence semantic similarity model \ie $\text{cosine}$(word objects, caption). Secondly, we compare our model against  two approaches that uses object information to improve image captioning: First, \citet{wang2018object} investigates  the benefit of object frequency counts for generating a good captions. We train an LSTM decoder (\ie language generation stage) with an object frequency counts dictionary on the training dataset. The dictionary is a Fully Connected layer, concatenated with the LSTM and a dense layer, that adds more weight to the most frequent  counts object that  are seen by the caption and the visual classifier. Second,  \citet{cornia2019show} that introduce a controllable grounded captions via a visual context. We train the last stage (decoder), attention and language model LSTM, on the training dataset to visually ground the generated caption based on the visual context.

One observation, shown in Table \ref{se:Table}, is that the benefit of using multiple visual contexts for longer captions, which can increase the chance of re-ranking the most visually related candidate caption, as shown in Figure \ref{fig:muti}  \simprob{} score with VilBERT.

%


Also, we investigate the statistical significance, using approximate randomization and bootstrapping resampling \cite{koehn2004statistical},  to detect minute differences in BLEU and METEOR, and NIST-BLEU\footnote{An improved version of BLEU rewarded infrequently used words by giving greater weighting to rarer words.}  \cite{doddington2002automatic} scores, as shown in Table \ref{tb:lexical diversity}, in which we observe the improvement  with our re-ranker over BLEU, METEOR\footnote{It has been previously observed that METEOR correlates better with human judgments than BLEU.} and NIST-BLEU. 
We would like to remark here that, with regards to subtle variations, the statistical significance of metrics such as BLEU, NIST-BLEU, and SacreBLEU \cite{post2018call} tend to disagree with human judgement \cite{mathur2020tangled,kocmi2021ship}. We therefore also conduct a human evaluation study (Section 6).

\begin{figure}[t!]
\centering 

\includegraphics[width=\columnwidth]{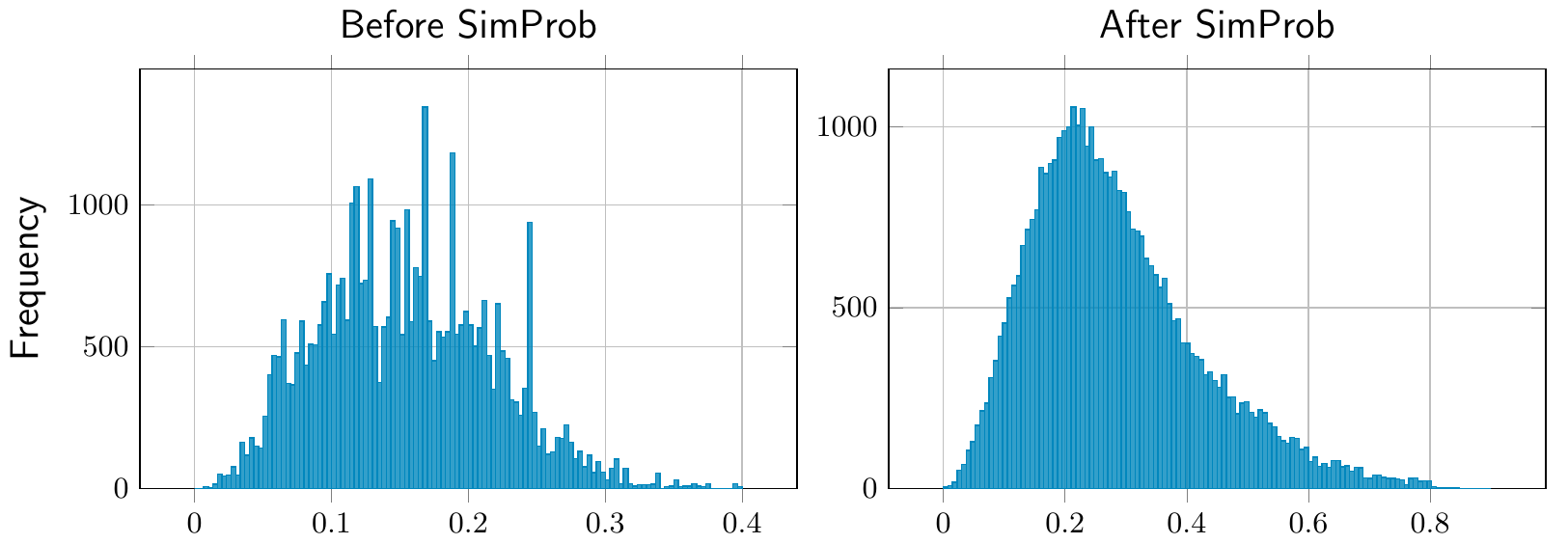}

\vspace{-0.2cm}

\caption{Visualization of the distribution change in re-rank scores on the 40k random sample from the test set. (Left) the score distribution before applying Belief Revision via \simprob{} with LM-GPT-2 initialization. (Right) the score distribution after applying the revision via similarity RoBERTa$_{\text{Large}}$ with the visual context.}

\label{fig:freq}
\end{figure}

Figure \ref{fig:freq} shows \simprob{} distribution over 40k samples with a pre-trained RoBERTa$_{\text{Large}}$ similarity score. (Left Figure) Before applying the revision, overall re-ranking scores are relatively low, and (Right Figure) after the visual revision, overall scores increased with more confident about each selected caption. The \simprob{} score positively shifts the distribution over all the samples.  


\subsection{Limitation}
We note that, the quality of the Beam search influences our re-ranker, since non-diverse, repeated captions or fewer novel ones, will make the re-ranking less effective, as shown in the Unique words per caption in Table \ref{tb:lexical diversity} with the Transformer baseline.

\begin{figure}[t!]
\small

\resizebox{\columnwidth}{!}{
 \begin{tabular}{W{3.3cm}  L{5cm}}
        \includegraphics[width=\linewidth, height = 2.5 cm]{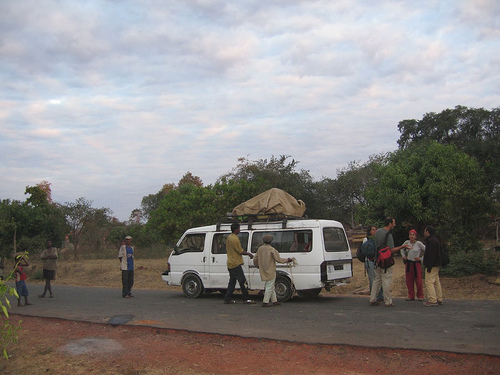} Visual: Minibus  &  \textbf{Vil$_{\text{BeamS}}$:}  a group of people standing around a white truck  \newline 
         \textbf{Vil+VR$_{\text{BERT}}$:}  a group of people standing around a white \textcolor{blue}{\textit{van}} \newline \textbf{Human:}  there is a white van that is stopped on the road  \\
     \arrayrulecolor{gray}\hline 
       \vspace{-0.2cm}
       \includegraphics[width=\linewidth, height = 2.5 cm]{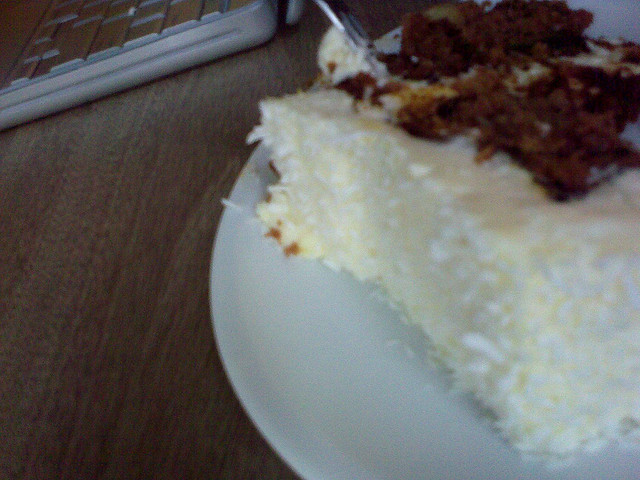} Visual: trifle & \textbf{Vil$_{\text{BeamS}}$:}  a close up of a plate of food  \newline 
         \textbf{Vil+VR$_{\text{BERT}}$:} \textcolor{blue}{\textit{piece}} of food sitting on top of \textcolor{blue}{\textit{a white plate}} \newline \textbf{Human:} a white plate and a piece of white cake  \\
         
         \arrayrulecolor{gray}\hline 
    \vspace{-0.2cm}
      \includegraphics[width=\linewidth, height = 2.5 cm]{COCO_val2014_000000202923.jpg} Visual: baseball  &  \textbf{Vil$_{\text{BeamS}}$:} a group of men on a field playing baseball  \newline 
         \textbf{Vil+VR$_{\text{BERT}}$:} \textcolor{blue}{\textit{a batter catcher and umpire during a baseball game}} \newline \textbf{Human:} batter catcher and umpire anticipating the next pitch \\
         
    \end{tabular}
}
 \vspace{-0.2cm}


\caption{Example of captions re-ranked by our Visual Re-ranker and the original caption (Best-beam) from the base system. Re-ranked captions are more precise, have a higher lexical diversity, or provide more details.}

\label{tb-fig:example}

\end{figure}

\section{Evaluation of Diversity}
We follow the standard diversity evaluation metrics \cite{shetty2017speaking,deshpande2019fast}: (1) \textit{Div-$1$} the ratio of unique unigram to the number of words in caption (2) \textit{Div-$2$} the ratio of unique bi-gram to the number of words in the caption, (3) \textit{mBLEU} is the BLEU score between the candidate caption against all human captions (lower value indicate diversity). However, since even though we obtained the top-20 candidates from the base systems, many of them are the same or have very small differences (beam search drawback), which will reflect in small performance differences before and after re-ranking. Therefore, some of the standard metrics are not able to capture these small changes, as shown in Table \ref{tb:lexical diversity}. Consequently, to try to capture the changes  and the effect of the re-ranking, we also measured the lexical and semantic diversity with the following metrics:

\noindent\textbf{Type-Token Ratio (TTR):} TTR \cite{brown2005encyclopedia} is the number of unique words or types divided by the total number of tokens in a text fragment.

\noindent{\bf{Measure of Textual Lexical Diversity (MTLD):}} MTLD \cite{mccarthy2010mtld} is based on TTR, and measures the average length of subsequences in the text for which a certain TTR is maintained, thus, unlike TTR, being length-invariant.

For semantic diversity, we use the standard metric \cite{wang2019describing} Self-CIDEr. Also, inspired by BERTscore and following \cite{song2021sentsim} that introduce Sentence Semantic  Semantic (SSS) for machine translation, we use SBERT-sts (fine-tuned on sts task) with a cosine score to measure the sentence level semantic correlation against all human references. We observe that the SBERT-sts capture the semantic content better than Self-CIDEr.

Table \ref{tb:lexical diversity} shows that visual re-ranking selects longer captions and with higher lexical diversity than the base system beam search. Figure \ref{tb-fig:example} shows some examples where visual context re-ranking selected captions with more precise lexica (\textit{van} vs. \textit{truck}), higher diversity --\ie adding details about objects (\textit{white plate} vs. \textit{plate})-- and even selecting a more specific abstraction level (\textit{batter, catcher and umpire} vs. \textit{a group of men}).

\begin{table*}[t!]\centering
\small 
\centering
\resizebox{\linewidth}{!}{
\begin{tabular}{lcccc|cc|cccccc|cccccc}\toprule 


& \multicolumn{4}{c}{Lexical Diversity} &   \multicolumn{2}{c}{Vocabulary} & \multicolumn{6}{c}{Accuracy 4-gram ($p$-value) } &  \multicolumn{2}{c}{ mBLEU$ \downarrow$  } & \multicolumn{2}{c}{$n$-gram Diversity} &  \multicolumn{2}{c}{Semantic Diversity}
\\\cmidrule(lr){2-5}\cmidrule(lr){6-7}\cmidrule(lr){8-13}\cmidrule(lr){14-15}\cmidrule(lr){16-17}\cmidrule(lr){18-19}

& MTLD & TTR &  Uniq & WPC&  Dist & Dist$^\ast$ & BLEU & $p$ &  M &  $p$ & NIST & $p$ & best-5   &  best Beam$^\ast$  & Div-1& Div-2  & Self-CIDEr &SBERT-sts\\

Human  & 19.56          &  0.90  &  9.14  & 14.5 & 3425  & 3326 \\  
 
\addlinespace[0.1cm]
\multicolumn{4}{l}{VilBERT}	\\ 
Vil$_{\text{BeamS}}$  &     17.28     	&  0.87	 &  8.05	&  10.5   &894 & 842 & 0.337 & - &   0.265  & - & 0.755 &- &  0.899& 0.454 &  0.38 	 & 0.44	   & 0.661 & 0.7550\\
Vil+VR$_{\text{w-Object}}$  \cite{fang2015captions}     & 15.90 &  0.87 	 & 8.02 & 9.20  &921  & 866 & 0.335 & 0.109 &  0.266 & 0.46 & 0.764 & 0.00 &0.899 & 0.455 & 0.38 	 & 0.44 & \textbf{0.662} & \textbf{0.7605} \\		
Vil+VR$_{\text{Object}}$ \cite{wang2018object}       & 15.77 &  0.87	 & 8.03 & 9.19  &911 &854 & 0.335  & 0.131 &0.266& 0.57 & 0.761 & 0.043& 0.899	 & 0.455 & 0.38  & 0.44 & 0.661 & 0.7570 \\	
Vil+VR$_{\text{Control}}$   \cite{cornia2019show}   & 15.69 & 0.87 	 & 8.07 & 9.21 & \textbf{935} & \textbf{878} & 0.331 & 0.016 & 0.266 & 0.46 &0.758  &0.118 & 0.899	 & 0.452 & 0.38  & 0.44 & 0.661 & 0.7567\\	
\rowcolor{Gray}
Vil+VR$_{\text{RoBERTa}}$ (ours)  (Table \ref{se:Table} Best)  & \textbf{17.70} &  0.87	 & \textbf{8.14} &\textbf{10.8} &   892 & 838& \textbf{0.339} & 0.147 & \textbf{0.267} & 0.04  & \textbf{0.764} & 0.002 &   \textbf{0.896}	& \textbf{0.451}  & 0.38 & 0.44 &  0.661 & 0.7562 \\	 
\midrule

\multicolumn{4}{l}{Transformer based caption generator} \\ 
Trans$_{\text{\text{BeamS}}}$   & 14.77 	& 	0.86  &7.44  & 9.62   & 935 & 897 & 0.341 & - &  0.272 & - & 0.781 &  -	& 	\textbf{0.954}  & 0.499 &  0.26 & 0.29 & 0.660 & 0.7707\\ 
Trans+VR$_{\text{w-Object}}$  \cite{fang2015captions}     & 13.14 &  0.85	 & 7.37 & 8.62 & 965 & 923 & 0.364 & 0.00   & 0.272 & 0.10 & 0.789 & 0.001 & 0.958  &   0.498 &  0.25 & 0.29 & 0.660 & 0.7709\\	
Trans+VR$_{\text{Object}}$ \cite{wang2018object}      & 13.38 &  0.86	 & \textbf{7.45} & 8.69  & \textbf{982}  & \textbf{940} & 0.369    & 0.00  & 0.271 & 0.04 & 0.798& 0.00 &  0.958 &    0.495 & 0.25 & 0.28& 0.660 & 0.7700 \\	
Trans+VR$_{\text{Control}}$   \cite{cornia2019show}    & 13.25 &  0.86	 & 7.44 & 8.64  & 961 & 921 &0.373 & 0.00  & 0.272 &  0.00 & 0.796& 0.00 & 0.958	 &     0.498 & 0.25 & 0.29 & 0.660 & \textbf{0.7716}\\	
\rowcolor{Gray}

Trans+VR$_{\text{BERT}}$ (ours) (Table \ref{se:Table} Best)  & \textbf{14.78}	&   0.86 & \textbf{7.45} & \textbf{9.76}  & 980 & 939 & \textbf{0.374}   & 0.00 &  \textbf{0.273}	& 0.19 &   \textbf{0.806} & 0.00 & 0.963 &      \textbf{0.338} & 0.26 & \textbf{0.30} & 0.660 &   0.7711 \\ 
\bottomrule 
\end{tabular}
}

\vspace{-0.2cm}
\caption{Diversity statistics and statistical tests. Measuring the diversity of caption before/after re-ranking. Uniq and WPC columns indicate the average of unique/total words per caption, respectively. The BLEU, \textbf{M}ETEOR and NIST are an average result,  with an approximate randomization  test with 1k trials, to estimate the statistically significant improvement  with/without  our re-ranker. mBLEU and mBLEU$^\ast$ are computed with respect to the top 5-captions and best-beam, respectively. We also report the Distinct vocabulary (Dist$^\ast$ filtering out stop words).} 
\label{tb:lexical diversity}
\end{table*}

\section{Human Evaluation}

We conducted a human study to investigate human preferences over the visual re-ranked caption. We randomly select 26 test images and give 12 reliable human subjects the option to choose between two captions:  (1)  Best-beam (BeamS)\footnote{The best result by the baseline in standard metrics.} and (2) \textbf{V}isual \textbf{R}-ranker. We obtain mixed results, as some re-ranked captions are grammatically incorrect, such as singulars instead of plurals and \textit{sitting on} for objects instead of subjects.  Overall, we can observe that 46\% of native speakers agreed with our visual re-ranker. Meanwhile, the result for non-native speakers is 61\%. In some details, we observe that our model and the non-native human subjects chose those re-ranked captions because they correlated more closely with the visual information regarding the grammatical error in the sentence and unlike the native speaker.




\section{Ablation study}
\vspace{-0.2cm}
Belief Revision relies on a different block (\ie LM, similarity and visual context) to make the final revision. In this study, we perform an ablation study over a random 100 samples from the test set to investigate the effectiveness of the proposed setup. Table \ref{tb:ablation} shows result with different settings. 




\vspace{0.1cm}
\noindent{\bf{Language Model Block:}} One of the principal intentions in initializing the original hypothesis with a LM is the ability to combine different models. We experimented with product probability, although the mean LM probability achieved better results.

\vspace{0.1cm}
\noindent{\bf{Similarity Block:}} The degree of similarity between the caption and its visual context is major factor in hypothesis revision. Thus, we experimented with a light model (Distil SBERT) and unsupervised/supervised \textbf{Sim}ple \textbf{C}ontrastive \textbf{S}entence \textbf{E}mbedding (SimCSE) for learning sentence similarity. The results show that unsupervised, via dropout with the sentence itself, contrastive learning based similarity performs well in the case of the longer captions, as shown in VilBERT Table \ref{tb:ablation}.


\begin{table}[t!]\centering

\small 
\resizebox{\columnwidth}{!}{
\begin{tabular}{lcccccc}


\toprule
 Model &  B-4 & M & R & C  & S \\
\midrule 


\multicolumn{6}{l}{VilBERT-VR-GPT-2$_{\text{mean}}$ + ResNet}	\\ 
\rowcolor{Gray}
+ RoBERTa  (Table \ref{se:Table} Best)   & 0.346 & 0.266 & \textbf{0.541} & \textbf{1.171} &   0.205 \\

+ LM-GPT-2$_{\text{product}}$  &    0.335     	&  0.266 &  0.535	&  1.142 & 0.205    \\


+ DistilSBERT   &     0.335    	&  0.266 &  0.537	&  1.128 &  0.205   \\

+ SimCSE \cite{gao2021simcse} &     0.324     	&   0.263 &   0.529	&  1.122 & \textbf{0.207}   \\
+ SimCSE (unsupervised) &     \textbf{0.349}     	&   \textbf{0.267} &  0.539	&  1.164 &  0.205   \\



+ CLIP \cite{radford2021learning} &    0.335     	&  0.261 &  0.527	&  1.142 &  0.202    \\

\midrule
\addlinespace[0.1cm]
\multicolumn{6}{l}{Trans - VR-GPT-2$_{\text{mean}}$ + ResNet}	\\ 
\rowcolor{Gray}
+ BERT (Table \ref{se:Table} Best)   & \textbf{0.363} & \textbf{0.268} & \textbf{0.565} & \textbf{1.281} &  0.207 \\


+ LM-GPT-2$_{\text{product}}$   &    0.360     	& 0.261 & 0.561	&  1.254 & 0.205    \\


+ DistilSBERT   &     0.355     	&  0.260 &  0.557	&  1.249 & 0.205   \\
+ SimCSE \cite{gao2021simcse} &     0.356     	&  0.265 &  0.564	&  1.272 & 0.207   \\
+ SimCSE (unsupervised) &    0.356     	&  0.263  &  0.560	&   1.253 & \textbf{0.208}  \\


+ CLIP \cite{radford2021learning} &    0.349     	&  0.260 &   0.555	&  1.243 & 0.203    \\

\bottomrule

\end{tabular}
}

\vspace{-0.2cm}
\caption{Ablation study using different information from various baselines in each block (\ie LM, similarity and visual context).}



\label{tb:ablation}
\end{table}

\begin{table*}[ht]
\small 
\centering
\resizebox{0.7\linewidth}{!}{
\begin{NiceTabular}{lccccccc}
\toprule   

 Model & B-1 &B-4 & M & R & C & S &BERTscore\\ 
\midrule
\multicolumn{8}{l}{VilBERT \cite{lu202012}}  \\
Vil$_{\text{Greedy}}$ & 0.751 &   0.330 & 0.272 & 0.554 &  1.104 & 0.207 &0.9352 \\ 


Vil$_{\text{BeamS}}$   & 0.752   &  0.351  &  0.274 &  0.557   & 1.115 & 0.205 & 0.9363 \\

\addlinespace[0.1cm]
Vil+VR$_{\text{W-Object}}$  \cite{fang2015captions}  & \textbf{0.756}  &  0.348  &  0.274 &  \textbf{0.559} & 1.123 &0.206 &  0.9365 \\ 
Vil+VR$_{\text{Object}}$  \cite{wang2018object}   & \textbf{0.756} &  0.348  &  0.274 &   \textbf{0.559}   & 1.120 & 0.206 & 0.9364 \\ 

Vil+VR$_{\text{Control}}$  \cite{cornia2019show}  & 0.753    &  0.345  &  0.274 &  0.557  & 1.116 & 0.206 & 0.9361 \\ 

\addlinespace[0.1cm]
\rowcolor{Gray} 
Vil+VR$_{\text{RoBERTa}}$  Table \ref{se:Table} (positive) & 0.753    &  \textbf{0.353} &  \textbf{0.276}    &  \textbf{0.559}   & 1.128  & 0.207 & \textbf{0.9366}  \\

\addlinespace[0.1cm]

Vil+VR$_{\text{RoBERTa}}^{-low}$  & 0.748    &  0.349  &  0.275 &  0.557  & 1.116 & 0.206  & 0.9362  \\ 


Vil+VR$_{\text{RoBERTa}}^{-high}$   & 0.748  &  0.349  &   0.275 &  0.557  & 1.116  & 0.206  & 0.9364  \\

Vil+VR$_{\text{GloVe}}^{-pos}$   & 0.751   &  0.351  &  0.276 &   0.558  & 1.123 & 0.207  & 0.9364  \\


\rowcolor{aliceblue}
    Vil+VR$_{\text{RoBERTa+GloVe}}^{-joint}$    &  0.750  &   0.351  &  \textbf{0.276} &   \textbf{0.559}  & \textbf{1.126} & \textbf{0.208}  &  \textbf{0.9365} \\

\midrule 
\addlinespace[0.1cm]
\multicolumn{8}{l}{Transformer based caption generator \cite{Marcella:20}}  \\
Trans$_{\text{Greedy}}$   & 0.787   & 0.368 & 0.276 & 0.574 &   1.211 & 0.215 & 0.9376 \\ 

Trans$_{\text{\text{BeamS}}}$ & 0.793   &   0.387  &0.281 &  0.582   &  1.247 & \textbf{0.220} & \textbf{0.9399}  \\ 

\addlinespace[0.1cm]
Vil+VR$_{\text{W-Object}}$  \cite{fang2015captions}  & 0.786  &  0.348  &  0.274 &  0.559 & 1.123 &0.206 &  0.9365 \\ 
Trans+VR$_{\text{Object}}$  \cite{wang2018object}   & 0.790   &  0.383 &   0.280 &0.580 &  1.237   & 0.219 &  0.9391  \\ 
Trans+VR$_{\text{Control}}$  \cite{cornia2019show} & 0.791   &   \textbf{0.388}   &  0.281 &  \textbf{0.583}  & 1.248 & \textbf{0.220} & 0.9398 \\ 

\addlinespace[0.1cm]
\rowcolor{Gray} 
Trans+VR$_{\text{BERT}}$ Table \ref{se:Table} (positive)   & 0.793  &	
 \textbf{0.388}   &  \textbf{0.282}  & \textbf{0.583} &   \textbf{1.250}  &  \textbf{0.220} & \textbf{0.9399}  \\  

Trans+VR$_{\text{BERT}}^{-low}$    & 0.791  &  0.387  &  0.280 &  0.582  & 1.242 & 0.218   &0.9396 \\
Trans+VR$_{\text{BERT}}^{-high}$   & 0.793   &  0.385  &  0.282 & 0.582  &  1.243 & 0.219  &  0.9397  \\



\rowcolor{aliceblue}

Trans+VR$_{\text{GloVe}}^{-pos}$  & \textbf{0.794}    &  \textbf{0.388}   &  \textbf{0.282} &  \textbf{0.583} & \textbf{1.249} & \textbf{0.220}  & \textbf{0.9399} \\

Trans+VR$_{\text{BERT+GloVe}}^{-joint}$  & 0.793 &  0.387  &  0.281 &  0.582  & 1.247 & \textbf{0.220}  & 0.9398  \\

\bottomrule

\end{NiceTabular}
}
\vspace{-0.2cm}
\caption{Comparison between positive (single concept VR) and Negative Belief Revision (NBR) on the Karpathy split. The NBR uses a \textit{high similarity} VR$^{-high}$ object related to the positive visual but not in the image, \textit{low similarity}  VR$^{-low}$ false positive from the visual classifier, and positive visual via static word level similarity VR$^{-pos}$. \textbf{Boldface} fonts reflect improvement over the baseline.}

\label{se:Table_2}
\end{table*}

\vspace{0.1cm}

\noindent{\bf{Visual Context Block:}} We experimented with the most recent model of Contrastive Language-Image Pre-Training (CLIP) \cite{radford2021learning} with Zero-Shot Prediction to extract the visual context. Although, CLIP can predict rare objects better, there is no improvement over ResNet152 with a huge computational  cost.

\section{Negative Evidence: an extension}


Until now, following \citet{blok2003probability}, we considered only the cases when the visual context increase the belief of the hypothesis (Equation~\ref{ch2: sim-to-prob-1}). \citet{blok2007induction} also propose Equation~\ref{eq:negative} for the case where the absence of evidence leads to a decrease in the probability of the hypothesis.   
\begin{equation}
\label{eq:negative}
\operatorname{P}(w \mid \neg c )=1-(1-\operatorname{P}(w))^{\alpha}
\end{equation}

In our case, we introduce negative evidence in three ways: 



\noindent{\textbf{False Positive Visual Context (VR$^{-low}$):}} We employ the false-positives produced by the visual classifier as negative information to decrease the hypotheses. In this case, we have lower similarity measures as the relation between the visual context and caption are farther apart.     
\vspace{0.1cm}

\noindent{\textbf{Absent Visual Context (VR$^{-high}$):}} The negative information here is a set of visual information extracted from the original visual context (\ie from the visual classifier) which does not exist in the image. Thus, the visual context produced by the classifier is used as a query on a pre-trained 840B GloVe \cite{pennington2014glove}, with cosine similarity, to retrieve the closest visual context in the same semantic space (\eg visual: river, closest visual: valley).

\vspace{0.1cm}

\vspace{0.1cm}

\noindent{\textbf{Positive Visual Context (VR$^{-pos}$):}} As the previous two-approaches produced unexpected results with low and high similarities as shown in Table \ref{se:Table_2},  we approach this from a positive belief revision perspective but as negative evidence. Until now, all approaches use sentence-level semantic similarity, but in this experiment, we convert the similarity from sentence to word level. For this  first, we employ an LSTM based CopyRNN keyphrase extractor \cite{meng2017deep}, which is trained on a combined pre-processed wikidump (\ie  keyword, short sentence) and SemEval 2017 Task 10 (Keyphrases from scientific publications) \cite{augenstein2017semeval}. Secondly, GloVe is used to compute the cosine similarity with the visual context in a word-level manner. We consider this as negative evidence for the following reasons: (1) the similarity is computed without the context of the sentence and (2) the static embedding is computed without knowing the sense of the word. The advantage of VR$^{-pos}$ is that a high similarity and confident visual information are present and thus satisfies the revision. 

\vspace{0.1cm}
\noindent{\textbf{Joint Evidence (VR$^{-joint}$):}} Finally, we combined the best model VR$^{-pos}$ with the best positive evidence (baseline+VR$_{\text{BERT/RoBERTa}}$) with a simple multiplication. 

Table~\ref{se:Table_2} shows that there is some refinement results with the negative evidence over both baselines with VR$^{-pos}$ and VR$^{-joint}$. However, there is no improvement over the original positive evidence.


\section{Discussion: Limitations}

\begin{figure}[t!]
\small

\resizebox{\columnwidth}{!}{
 \begin{tabular}{W{3.3cm}  L{5cm}}
        \includegraphics[width=\linewidth, height = 2.5 cm]{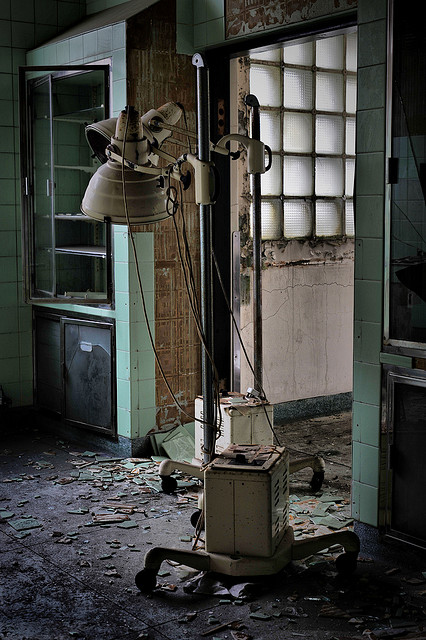} Visual: vacuum {\color{red}{\xmark}} & 
       \textbf{Vil$_{\text{BeamS}}$:} a pile of trash sitting inside of a building
      \newline   
      \textbf{Vil+VR$_{\text{BERT}}$:}  a pile of trash sitting in front of a building {\color{red}{\xmark}}  
      \newline \textbf{Human:} an older floor light sits deserted in an abandoned hospital \\

     \arrayrulecolor{gray}\hline 

    \vspace{-0.2cm}
      \includegraphics[width=\linewidth, height = 2.5 cm]{COCO_val2014_000000293611.jpg}  Visual: barbershop {\color{red}{\xmark}}&   
  \newline \textbf{Trans$_{\text{BeamS}}$:} a kitchen with black counter tops  and wooden cabinets 
          \newline
      \textbf{Tans+VR$_{\text{BERT}}$:}  a kitchen counter with a black  counter top  {\color{red}{\xmark}} 
      \newline \textbf{Human:} a kitchen with a sink bottles jars and a dishwasher\\
   
    \end{tabular}
}
 \vspace{-0.2cm}
 \caption{Failure cases of the object detectors. The object classifier struggle with images with complex background and out-of-context object. }

\label{tb-fig:limitation_paper}
\end{figure}

\noindent{\bf{Object Classifier Failure Cases:}} As the belief revision approach relies heavily on the object in the image for the likelihood revision, the quality of the object classifier is critical for the final decision. Here, we show  some failure cases when the visual classifier struggle with complex background (\ie wrong visual, object hallucination, \etc) as shown in Figure \ref{tb-fig:limitation_paper}. Note that, if no related visual is present in the image the belief revision score will back off to 1 (no revision needed).

\noindent{\bf{Object-to-Caption Similarity Score:}} Another limitation is the low/high cosine similarity score, which unbalances the likelihood revision. For example, a visual context \textit{paddle} and the caption: \textit{a man riding a surfboard on a wave} have low cosine scores when using a pre-trained model that is fine-tuned on sentence-to-sentence semantic similarity tasks (\ie STS-B).  Note that, we tackle this problem by adding multiple visual contexts as shown in Figure \ref{fig:muti}. 





\section*{Related work}

Modern sophisticated image captioning systems focus heavily on visual grounding to capture real-world scenarios.  Early work \citet{fang2015captions} builds a visual detector to guide and re-ranked image captioning with global similarity. The work of \cite{wang2018object} investigates the informativeness of visual or object information (\eg object frequency count, size and position) in an end-to-end caption generation. Another work \citet{cornia2019show} proposes controlled caption language grounding through visual regions from the image. More recently, \citet{gupta2020contrastive} introduce weakly supervised contrastive learning via object context and language modeling (\ie BERT) for caption phrase grounding. 
%
%
%
Inspired by these works, \cite{fang2015captions} carried out re-ranking via visual information, \cite{wang2018object,cornia2019show, chen2020say}  explored the benefits of object information in image captioning, \cite{gupta2020contrastive} exploited the  benefits of language modeling to extract contextualized word representations and the exploitation of the semantic coherency in caption language grounding  \cite{zhang2021consensus}, we purpose an object based re-ranker via human inspired logic reasoning with Belief Revision to re-rank the most closely related captions with contextualized semantic similarity.  
%

Unlike the earlier approaches, our methods employ state-of-the-art tools and pre-trained models. Therefore, the system will keep improving in the future as better systems become available. In addition, our model can be directly used as a drop-in complement for any caption system that outputs a list of candidate hypothesis.


%



\section*{Conclusion}
In this work, we aim to demonstrate that the Belief Revision approach that works well with human judgment can be applied to Image Captioning by employing human-inspired reasoning via a pre-trained model (\ie GPT, BERT). Belief Revision (BR) is an approach for obtaining the likelihood revisions based on similarity scores via human judgment. We demonstrate the benefits of the approach by showing that two state-of-the-art Transformer-based image captioning results are improved via simple language grounding with visual context information. In particular, we show the accuracy gain in a benchmark dataset using two methods: (1) BR with positive visual evidence  (increase the hypothesis)  and (2) negative evidence (decrease the hypothesis), with wrong visual \ie false positive by the classifier. However, this adaptation  could be applied to many re-ranking tasks in NLP (text generation, multimodel MT, lexical selection, \etc), as well as in Computer Vision  applications such as visual storytelling.

\section*{Ethical Considerations}

The core contribution of our paper is algorithmic for the task of image captioning. As such we do not foresee any downstream harms propagated immediately by our proposal. We however acknowledge that due to the very nature of the data driven processing, there could be an amplification or propagation of potential biases existing in the datasets.

\bibliography{anthology,custom}
\bibliographystyle{acl_natbib}
\appendix
\section{Appendix: Additional Result}
\label{sec:appendix}

\begin{table*}[t!]
\centering
\resizebox{\textwidth}{!}{
\begin{tabular}{lcccccccc}


& \multicolumn{4}{c}{BLEU ($p$-value)} & \multicolumn{4}{c}{NIST ($p$-value)} \\ 
 \cmidrule(lr){2-5}\cmidrule(lr){6-9}
Model & B1 & B2  & B3 & B4 & N1 & N2 & N3 & N4 \\

\multicolumn{4}{l}{VilBERT}	\\ 
Vil$_{\text{BeamS}}$   & 0.740 \small{\textcolor{white}{{(0.019)}}} & 0.578 \small{\textcolor{white}{{(0.019)}}}  & 0.441 \small{\textcolor{white}{{(0.019)}}}  & 0.337  \small{\textcolor{white}{{(0.019)}}} & 0.492 \small{\textcolor{white}{{(0.019)}}} & 0.672 \small{\textcolor{white}{{(0.019)}}} & 0.731 \small{\textcolor{white}{{(0.019)}}} & 0.755 \small{\textcolor{white}{{(0.019)}}} \\

Vil+VR$_{\text{w-Object}}$  \cite{fang2015captions}  & 0.744 \small{(0.019)} & \textbf{0.581} \small{(0.063)}  & 0.442 \small{(0.371)} & 0.335 \small{(0.109)} & \textbf{0.497} \small{(0.00) \textcolor{white}{{0}}} & \textbf{0.680} \small{(0.00)\textcolor{white}{{0}}} &  \textbf{0.740} \small{(0.001)} & 0.764 \small{(0.00)\textcolor{white}{{0}}}   \\  
Vil+VR$_{\text{Object}}$ \cite{wang2018object}    & \textbf{0.745} \small{(0.01)\textcolor{white}{{0}}} & \textbf{0.581} \small{(0.091)} & 0.441 \small{(0.429)} & 0.335  \small{(0.131)} & 0.496 \small{(0.003)} & 0.677 \small{(0.015)}& 0.737 \small{(0.012)} & 0.761 \small{(0.043)}  \\ 
Vil+VR$_{\text{Control}}$   \cite{cornia2019show}   & 0.741 \small{(0.233)} & 0.577 \small{(0.29)\textcolor{white}{{0}}}  &  0.439 \small{(0.104)} & 0.331 \small{(0.016)} & 0.494 \small{(0.015)} & 0.676 \small{(0.046)}&  0.735 \small{(0.064)} &  0.758 \small{(0.118)} \\
\rowcolor{Gray}  
Vil+VR$_{\text{RoBERTa}}$ (ours) (Table \ref{se:Table} Best)  & 0.741 \small{(0.382)} & 0.580 \small{(0.129)} &  \textbf{0.443} \small{(0.00)\textcolor{Gray}{{0}}} & \textbf{0.339} \small{(0.147)}  & \textbf{0.497} \small{(0.00)\textcolor{Gray}{{0}}} & \textbf{0.680}  \small{(0.00)\textcolor{Gray}{{0}}} & 0.740 \small{(0.00)\textcolor{Gray}{{0}}} & \textbf{0.764} \small{(0.002)} \\ 
\midrule


\multicolumn{4}{l}{Transformer based caption generator} \\ 
Trans$_{\text{\text{BeamS}}}$  & 0.726  \small{\textcolor{white}{{(0.019)}}}  & 0.584  \small{\textcolor{white}{{(0.019)}}}  & 0.451  \small{\textcolor{white}{{(0.019)}}} &  0.341 \small{\textcolor{white}{{(0.019)}}}  & 0.492 \small{\textcolor{white}{{(0.019)}}}  & 0.688 \small{\textcolor{white}{{(0.019)}}} & 0.756 \small{\textcolor{white}{{(0.019)}}} & 0.781 \small{\textcolor{white}{{(0.019)}}}    \\	  
Trans+VR$_{\text{w-Object}}$  \cite{fang2015captions}     & 0.775 \small{(0.00)\textcolor{white}{{0}}} & 0.625 \small{(0.00)\textcolor{white}{{0}}}  & 0.482 \small{(0.00)\textcolor{white}{{0}}} & 0.364 \small{(0.00)\textcolor{Gray}{{0}}}  & 0.498 \small{(0.00)\textcolor{white}{{0}}} & 0.696 \small{(0.00)\textcolor{white}{{0}}}  & 0.764 \small{(0.001)} & 0.789  \small{(0.001)}      \\	
Trans+VR$_{\text{Object}}$ \cite{wang2018object}     & 0.777 \small{(0.00)\textcolor{white}{{0}}} &0.628  \small{(0.00)\textcolor{white}{{0}}}  & 0.486 \small{(0.00)\textcolor{white}{{0}}}  & 0.369 \small{(0.00)\textcolor{white}{{0}}}& 0.504 \small{(0.00)\textcolor{white}{{0}}} & 0.703 \small{(0.00)\textcolor{white}{{0}}} & 0.773 \small{(0.00)\textcolor{white}{{0}}}  & 0.798 \small{(0.00)\textcolor{white}{{0}}}    \\	
Trans+VR$_{\text{Control}}$   \cite{cornia2019show}  & 0.780 \small{(0.00)\textcolor{white}{{0}}}  & 0.623  \small{(0.00)\textcolor{white}{{0}}}  & \textbf{0.490}  \small{(0.00)\textcolor{white}{{0}}}  & 0.373   \small{(0.00)\textcolor{white}{{0}}} & 0.502 \small{(0.00)\textcolor{white}{{0}}}  & 0.700 \small{(0.00)\textcolor{white}{{0}}} & 0.771 \small{(0.00)\textcolor{white}{{0}}}  &  0.796 \small{(0.00)\textcolor{white}{{0}}}     \\	
\rowcolor{Gray}

Trans+VR$_{\text{BERT}}$  (ours) (Table \ref{se:Table} Best)  & \textbf{0.781} \small{(0.00)\textcolor{Gray}{{0}}} & \textbf{0.631} \small{(0.00)\textcolor{Gray}{{0}}}  & \textbf{0.490} \small{(0.00)\textcolor{Gray}{{0}}}  & \textbf{0.374} \small{(0.00)\textcolor{Gray}{{0}}}  & \textbf{0.509} \small{(0.00)\textcolor{Gray}{{0}}} & \textbf{0.710} \small{(0.00)\textcolor{Gray}{{0}}} & \textbf{0.781} \small{(0.00)\textcolor{Gray}{{0}}}  & \textbf{0.806}  \small{(0.00)\textcolor{Gray}{{0}}}   \\

\hline 

\end{tabular}
}
\vspace{-0.2cm}
\caption{Result with pair bootstrapping resampling  test via 1k trial \cite{koehn2004statistical} on the significant improvement before and ranking with BLEU and NIST. }
\label{tb:bootst}
\end{table*}

\begin{table*}[t]
\small 
\centering

\vspace{0.1cm}
\resizebox{\textwidth}{!}{
\begin{tabular}{lccccc}

 & \multicolumn{5}{c}{SacreBLEU} \\ 
\cmidrule(lr){2-6} 
 Model & Baseline Avg & New Avg & delta & Baseline better confidence \% & New better confidence \% \\

\multicolumn{5}{l}{VilBERT \cite{lu202012}} \\

Vil$_{\text{BeamS}}$  & 9.10&   &  &  & \\
\addlinespace[0.1cm]

Vil+VR$_{\text{W-Object}}$  \cite{fang2015captions}  & &  9.18 & 0.08 & 27.60 & 72.40 \\ 
Vil+VR$_{\text{Object}}$   \cite{wang2018object}    &  & 8.89 & -0.22 & 93.50 & 6.50 \\
Vil+VR$_{\text{Control}}$  \cite{cornia2019show}  & & 9.01 & -0.09 & 75.60 & 24.40 \\


\rowcolor{Gray} 
Vil+VR$_{\text{RoBERTa}}$ (ours)  (Table \ref{se:Table} Best)  & & \textbf{9.29} & 0.18 & 8.50 & \textbf{91.50} \\

\hline 

\multicolumn{5}{l}{Transformer Caption Generator \cite{Marcella:20}} \\ 

\addlinespace[0.1cm]

Trans$_{\text{\text{BeamS}}}$   & 10.16 &  &  &  &  \\ 

\addlinespace[0.1cm]
Trans+VR$_{\text{W-Object}}$  \cite{fang2015captions} & & 10.01 & -0.16 & 93.80  & 6.20 \\

Trans+VR$_{\text{Object}}$   \cite{wang2018object}  & & 10.18 & 0.02 & 43.40 & 56.60 \\ 

Trans+VR$_{\text{Control}}$  \cite{cornia2019show}   & &  10.09 & -0.07 & 83.80 & 16.20 \\

\rowcolor{Gray} 
Trans+VR$_{\text{BERT}}$ (ours) (Table \ref{se:Table} Best)  &  & \textbf{10.36} & 0.19 & 1.10 & \textbf{98.90} \\

\hline 
\end{tabular}
}
\caption{Result with pair bootstrapping resampling  test via 1k trial \cite{koehn2004statistical} on the significant improvement before and ranking with Sacrebleu \cite{post2018call}.}


\label{se:scareBLEU}
\end{table*}

\noindent{\textbf{\simprob{} with Two Visual Contexts.}}
Until now, we experimented with one  or multiple visual contexts at the same time; however, when we have more than one evidence supporting the revision, we can reason and choose which one to use to revise the hypothesis. In this scenario we begin with the conditional probability that comes from the dominant premise alone $P(Q_c | Q_a)$ (let's assume $Q_a$ is dominant). Then we add a fraction of the remaining lack of trust or confidence  $1 - P(Q_c|Q_a)$ that the dominant conditional \textit{leaves behind}. The similarity between  premise  categories defines the size of the portion or fraction  $sim(a,b)$ and separates the influence of the nondominant premise on the conclusion prior  $P(Q_c|Q_b)-P(Q_c)$. The component of similarity is designed to reduce the impact of the non-dominant premise when the premises are redundant. Note that the proposed method in equation below guarantees an increase in strength with additional premises. In addition, this property,  $\operatorname{Pr}(Q c | Q a, Q b) \geq \operatorname{Pr}(Q c | Q a), \operatorname{Pr}(Q c | Q b)$, is noncompetitive in the sense that one category does not reduce the probability of concept for another. Following the notation of Equation \ref{ch2: sim-to-prob-1}, we write the two visual contexts \simprob{} as:

$\small \operatorname{P}(w \mid c_{1}, c_{2})=\beta M+(1-\beta) S$, where
$$
\small 
\begin{aligned}
&\beta=\max \left\{\begin{array}{l}
\operatorname{sim}(w,  c_{1}) \\
\operatorname{sim}(w,  c_{2}) \\
\operatorname{sim}(c_{1}, c_{2}) \\
1.0-\operatorname{sim}(w,  c_{1}) \\
1.0-\operatorname{sim}(w,  c_{2}) \\
\operatorname{P}(c_{1}) \\
\operatorname{P}(c_{2})
\end{array}\right\} \\
&M=\max \{\operatorname{P}(w \mid c_{1}), \operatorname{P}(w \mid c_{2}\}, S= \\
&\operatorname{P}(w \mid c_{1})+\operatorname{P}(w \mid c_{2})-\operatorname{P}(w \mid c_{1}) \times \operatorname{P}(w \mid c_{2})
\end{aligned}
$$
\noindent where $\operatorname{P}(w \mid c_{1})$ and $\operatorname{P}(w \mid c_{2})$ are defined by Equation 1, and the two visual contexts are: (1) $c_{1}$  ResNet is the label \textbf{Class} and  (2) the $c_{2}$ COCO \textbf{Object} categories are from Inception-ResNet v2 based Faster RCNN. Note that $\beta$ takes the $max$ of all models, and thus it is not breaking the formation if one of the similarities or probabilities  is not confident enough (\ie if it is below the threshold).

 The last rows in Table \ref{se:Table_app} show the result of the \simprob{ } selecting the best visual context of the two visuals. However, although there is some improvement over the other approaches (\ie single and multi-visual contexts),  it is not significant enough to justify  the computational cost as a post-processing approach.

\begin{figure}[t!]
\centering 

\includegraphics[width=\columnwidth]{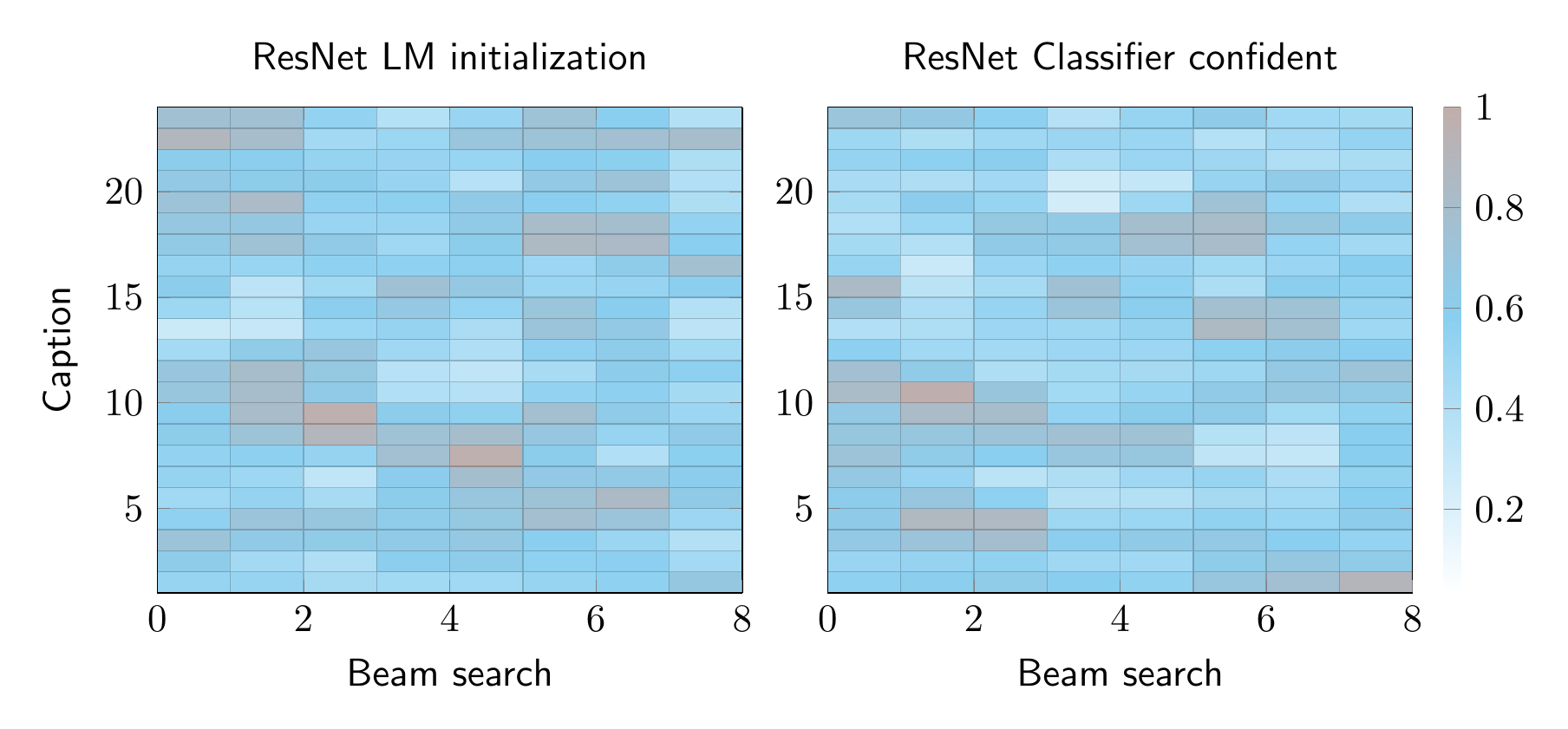}

\vspace{-0.3cm}

\caption{\simprob{} score of top-8 Beam search caption re-ranking (Right) with the visual classifier confidence probability without any initialization, (Left) with visual context that initialized by general common observation \ie LM. }  
\label{fig:init}
\end{figure}

Figure \ref{fig:init} shows the benefit of employing common observation via Unigram LM in comparison to the classifier confident, (Left Figure) a denser \simprob{} score caption re-ranking.

\noindent{\textbf{Additional Statistical Significance Analysis.}} 
Table \ref{tb:bootst} shows the full results of the statistically significant test via pair bootstrap resampling  \cite{koehn2004statistical}\footnote{\url{https://github.com/moses-smt/mosesdecoder/tree/master/scripts/analysis}} with BLEU and NIST. The NIST metric is an improved version of BELU that rewards infrequently used words by giving greater weighting to rarer words. 

Also, we employ Sacrebleu\footnote{\url{https://github.com/pytorch/translate/blob/master/pytorch_translate/bleu_significance.py}} \cite{post2018call} to investigate the statistically significant improvement, using delta, of our re-ranker with a BLEU score using the same approach as that above. Table \ref{se:scareBLEU} shows that our method performs better as \textbf{new better confidence} than the two baselines.  


\noindent{\textbf{Additional Diversity Analysis.}} Table \ref{tb:POS} shows part-of-speech tagging (POS) results before and after visual re-ranking. The proposed model VR yields a richer output in all POS tags in both baselines.



\vspace{0.1cm}

\noindent{\textbf{Full Experimental Results.}} Table \ref{se:Table_app} and Table \ref{se:Table_neg} show the full results of our experiments, with the most common metrics used for image captioning, for positive and negative evidence, respectively. Also, as we mentioned before, inspired by BERTscore and following \cite{song2021sentsim} we employ sentence-to-sentence semantic similarity score to compare candidate captions with human references with pre-trained Sentence-RoBERTa$_{\text{LARGE}}$ \cite{reimers2019sentence} tuned for general STS task. Unlike BERTscore which aligns word-to-word similarites, SBERT-sts builds a semantic vector for the whole sentence, which can be used to compare candidate captions with human references.
\vspace{0.1cm}

\noindent{\textbf{Additional Ablation Study.}} Table \ref{tb:comparsion} shows additional ablation study experiments with different information. 


\begin{figure}[h!]

\centering 

\includegraphics[width=0.6\columnwidth]{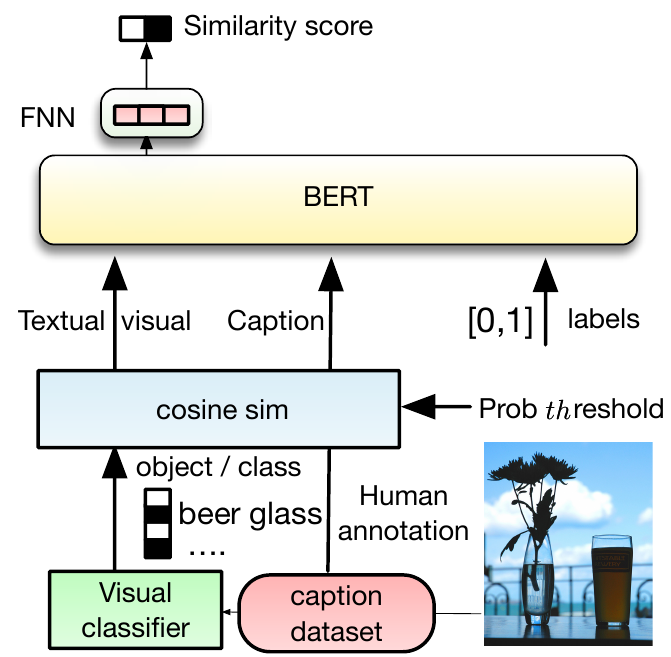}

\vspace{-0.1cm}

\caption{Dataset preprocessing until training.  We use two methods to filter our non-related visual context  (1) probability threshold: to filter out the visual context, and (2) semantic alignment with the caption via cosine distance (semantic relation).}

\label{fig:data-c}
\end{figure}

\noindent{\textbf{Training Dataset.}} As shown in Figure \ref{fig:data-c}, we use two approaches to match and filter out not related visual context: 1) Threshold: to filter out the probabilities prediction when the visual classifier is not confident. 2) Semantic alignment: to match the most related caption to its environmental context. In more detail,  we use cosine similarity with GloVe to match the visual with its context.  Table \ref{tb:dataset-coco}  illustrates samples of the enriched human annotation, and caption dataset, with visual context information.


\begin{table}[h!]
\centering
\small

\begin{tabular}{lcccc}

\hline 

Model & Noun & Verb & Adj & Conj \\ 
    \hline 
     \multicolumn{5}{l}{VilBERT} \\

Vil$_{\text{BeamS}}$              & 14094 & 3586  &3220    &  7914   \\ 
Vil+VR$_{\text{RoBERTa}}$  & \textbf{14403}   & \textbf{3739}  & \textbf{3325} &\textbf{8233}     \\ 

\hline 
 \multicolumn{3}{l}{Transformer} \\

Trans$_{\text{BeamS}}$              &  13961    & 3111  & 2004    & 7458   \\ 
Trans+VR$_{\text{BERT}}$  & \textbf{14203}  &\textbf{3146}  &\textbf{2056}   & \textbf{7563}   \\

\hline

\end{tabular}

\caption{Most frequent POS tag before and after visual re-ranking. The result shows that after Visual Re-ranking both captions have more noun, verb, \etc}

\label{tb:POS}
\end{table}

\begin{figure}[!]

\centering 

\includegraphics[width=0.9\columnwidth]{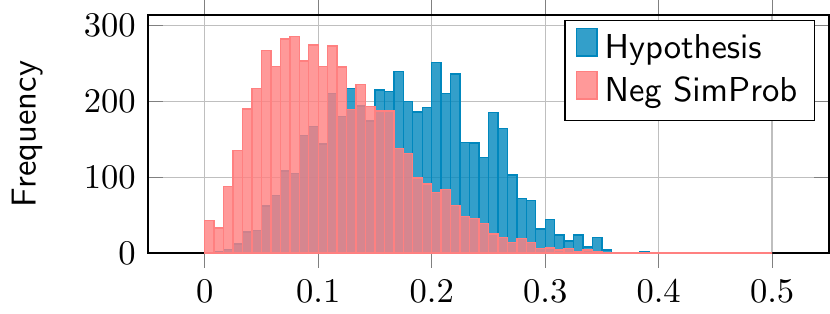}
\vspace{-0.2cm}

\caption{Visualization of Negative Evidence Neg-\simprob{} distribution on random samples from the test set (karpathy split). The negative visual information VR$^{-pos}$ decreases the hypothesis that is initialized by LM-GPT-2. }


\label{fig:negtive}
\end{figure}

\noindent{\textbf{Why Positive Visual as Negative  Evidence ?}} As we mentioned in the main script, we consider this as negative evidence for the following reasons: (1) the similarity is computed without the context of the sentence and (2) the static embedding (without knowing the sense of the word, \eg \textit{bar} for alcoholic drinking or rectangular solid piece of block). Although this approach relies on positive information, which is not the main intention of the negative evidence, the results demonstrate that there is a new direction of research that can be conducted using positive information as negative evidence. Figure \ref{fig:negtive} shows that the original hypothesis is decreased with the positive information. 


\begin{figure}[h!]
\centering 
\includegraphics[width=0.8\columnwidth]{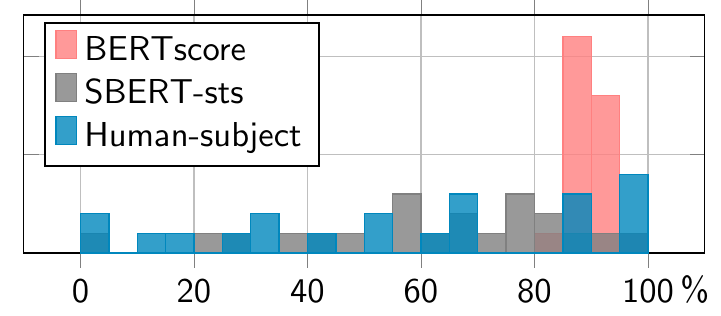}

\vspace{-0.4cm}

 \caption{Comparison results between native human subject, BERTscore, and sentence level metric SBERT-sts on the test set.}

\label{fig:human}

\end{figure} 

\begin{figure}[ht!]
 \small 
\centering 

\includegraphics[width=5.8cm]{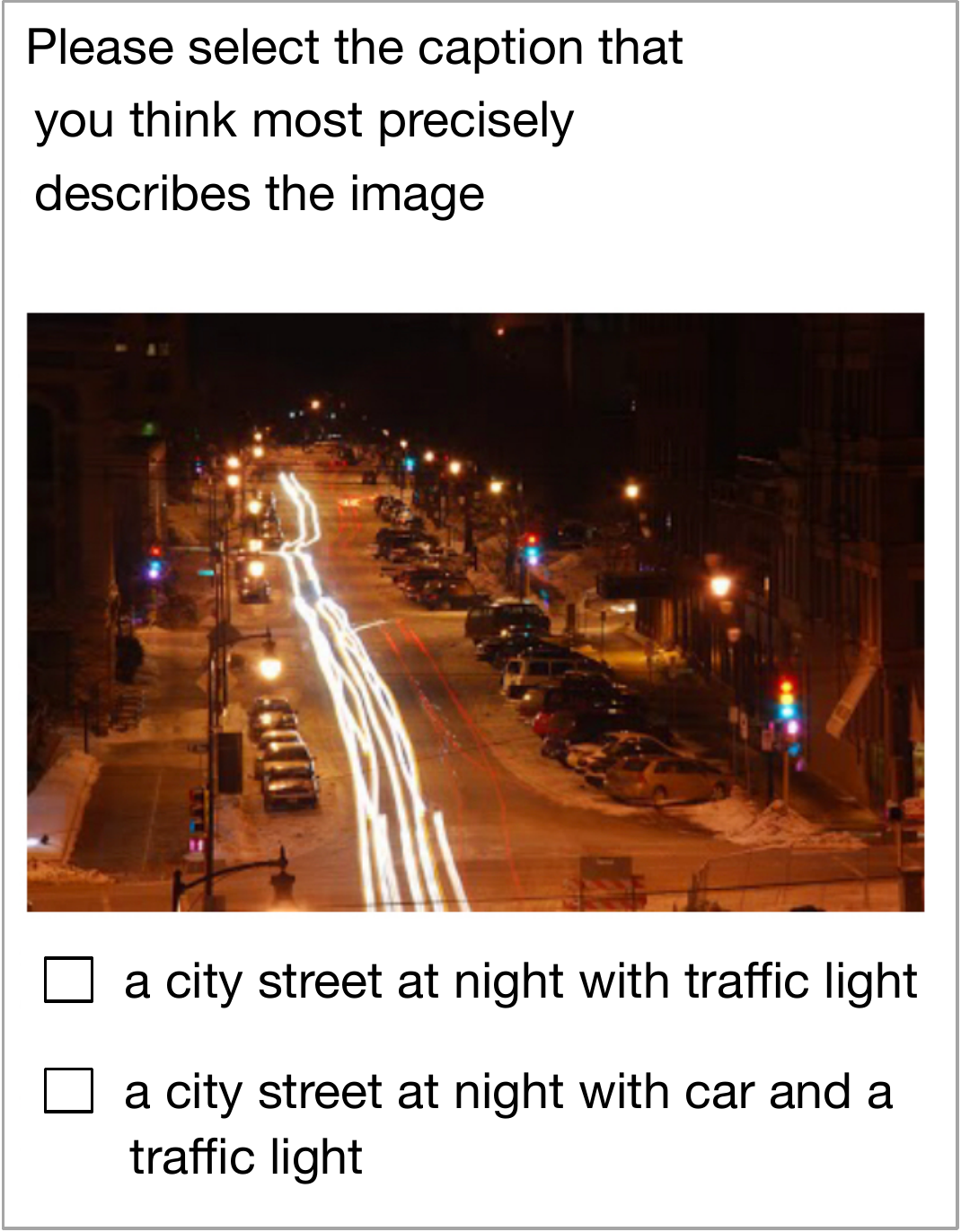}
\caption{\textbf{Human Evaluation}. The user interface presented to our human subjects through the survey
website asking them to re-rank the most descriptive caption candidates based on the visual information.}

\label{fig:HR}
 \end{figure}

\begin{table*}[t]
\small 
\centering

\vspace{0.1cm}
\resizebox{\textwidth}{!}{
\begin{tabular}{l c c c c c ccccc}

 \hline 
Model & B-1  & B-2  & B-3 &  B-4 &  M &  R  & C & S & BERTscore& SBERT-sts \\
 \hline



\multicolumn{11}{l}{VilBERT \cite{lu202012}}\\ 

Vil$_{\text{Greedy}}$     & 0.751  & 0.587  & 0.441 & 0.330 & 0.272 & 0.554 &  1.104 & 0.207 &0.9352 & 0.7550\\ 
\vspace{0.1cm}

Vil$_{\text{BeamS}}$     & 0.752 &  0.592  & 0.456    &  0.351  &  0.274 &  0.557   & 1.115 & 0.205 & 0.9363 & 0.7550 \\ 
\vspace{0.1cm}


Vil+VR$_{\text{W-Object}}$  \cite{fang2015captions} & 0.756 &  0.595  & 0.456    &  0.348  &  0.274 & 0.559 & 1.123 &0.206 &  0.9365 &  \textbf{0.7605} \\ 
Vil+VR$_{\text{Object}}$   \cite{wang2018object}   & 0.756 &  0.594  & 0.455    &  0.348  &  0.274 &  0.559   & 1.120 & 0.206 & 0.9364 &0.7570 \\


Vil+VR$_{\text{Control}}$  \cite{cornia2019show}   & 0.753 &  0.591  & 0.453    &  0.345  &  0.274 &  0.557  & 1.116 & 0.206 & 0.9361 & 0.7565 \\

\vspace{0.1cm}


Vil+GPT-2$_{\text{mean}}$ (only LM) &  0.749 &  0.590 & 0.455 & 0.351 &  0.276 & 0.558 & 1.124 & 0.208 & 0.9364 & 0.7546 \\
Vil+VR$_{\text{BERT}}$  (only $sim$) &  0.753   &  0.591 &   0.452  &  0.343    &  0.273  & 0.556  & 1.112 & 0.206 & 0.9361 & 0.7562 \\ 
 \hline
Vil+VR$_{\text{BERT}}$   & 0.752 &   0.592 &   0.456   &  0.351 &  0.274    &  0.557  & 1.115 & 0.205 & 0.9365 & 0.7567\\
Vil+VR$_{\text{BERT-Object}}$  &  0.752  &   0.592   &  0.457 &   0.352  &   0.277 & 0.560 &   1.129  & 0.208 & 0.9365   & 0.7562\\



Vil+VR$_{\text{RoBERTa}}$     & 0.753 & 0.594  & 0.458   &  \textbf{0.353} &  0.276 & 0.559   & 1.128  & 0.207 & \textbf{0.9366}  &  0.7562 \\

Vil+VR$_{\text{RoBERTa-Object}}$  &  \textbf{0.758} &  \textbf{0.611} & \textbf{0.465} &  0.344 & 0.262  &  0.555 &  1.234 &  0.206 & 0.9365 & 0.7554  \\

\hline 
Vil+VR$_{\text{BERT-Multi-class}}$   & 0.753 &  0.593 & 0.458 &   \textbf{0.353}  &  0.276    &  0.559  &  1.131 & 0.208  & 0.9365 &  0.7586\\
Vil+VR$_{\text{BERT-Multi-object}}$   &  0.752 & 0.592 & 0.456 &   0.351 &  0.276    &  0.558  &  1.123 &  0.208  & 0.9364 & 0.7566\\
 \rowcolor{Gray}
Vil+VR$_{\text{RoBERTa-Multi-class}}$  & 0.751 & 0.591 & 0.456 &  0.351 &   \textbf{0.277}    &  \textbf{0.561}   & \textbf{1.137}   & 0.208 &   \textbf{0.9366}   & 0.7589\\ 


Vil+VR$_{\text{RoBERTa-Multi-object}}$  & 0.753 & 0.593 & 0.458 &   \textbf{0.353} &   0.276    &  0.559   & 1.131  & 0.208 &   0.9365  & 0.7586 \\ 
\hline

Vil+VR$_{\text{BERT-Class+object}}$  & 0.752 &0.592 &0.455 & 0.350  &  0.276 &   0.559 & 1.126  & \textbf{0.209} &   0.9365  & 0.7563 \\


Vil+VR$_{\text{RoBERTa-class+object}}$  & 0.752& 0.592 &0.457 & 0.352  & \textbf{0.277} &  0.559 &1.127  & 0.208 &   0.9365  & 0.7558 \\ 
 


\hline 

\multicolumn{11}{l}{Transformer Caption Generator \cite{Marcella:20}}\\ 
Trans$_{\text{Greedy}}$     & 0.787   & 0.634 & 0.488 & 0.368 & 0.276 & 0.574 &   1.211 & 0.215 & 0.9376 & 0.7649\\ 
\vspace{0.1cm}
Trans$_{\text{\text{BeamS}}}$      &  0.793 &  0.645  & 0.504 &   0.387  &0.281 &  0.582   &  1.247 & \textbf{0.220} & \textbf{0.9399} & 0.7707 \\ 
Trans+VR$_{\text{W-Object}}$  \cite{fang2015captions} &  0.786 &   0.638  & 0.497    &  0.378  &  0.277 &   0.579 & 1.228 & 0.216 &  0.9388 & 0.7709 \\

Trans+VR$_{\text{Object}}$   \cite{wang2018object}   &  0.790 &   0.642  &  0.501    &  0.383 &   0.280 &0.580 &  1.237   & 0.219 &  0.9391 & 0.7700  \\ 


Trans+VR$_{\text{Control}}$  \cite{cornia2019show}  & 0.791 &  0.644  & \textbf{0.505}    &  \textbf{0.388}  &  0.281 &  \textbf{0.583}  & 1.248 & \textbf{0.220} & 0.9398 & 0.7716 \\ 

 \vspace{0.1cm}
 Trans+GPT-2$_{\text{mean}}$ (only LM)&  0.791 &  0.643 & 0.503 & 0.386 &  0.281 & 0.582 & 1.242 & 0.219 & 0.9396 &  0.7714\\

Trans+VR$_{\text{BERT}}$ (only $sim$)   &  0.789   & 0.640 &  0.498  & 0.380   &  0.279  & 0.579 & 1.234 & 0.219 & 0.9389 & 0.7693 \\ 
 \hline 
 
\rowcolor{Gray}

Trans+VR$_{\text{BERT}}$  &  0.793  &  \textbf{0.646}   &  \textbf{0.505} &  \textbf{0.388}   &  \textbf{0.282}  & \textbf{0.583} &   1.250  &  \textbf{0.220} & \textbf{0.9399}  & 0.7711 \\  

Trans+VR$_{\text{BERT-Object}}$  & 0.793 &   0.644   &  0.503 &   0.385  &   0.281 &  0.581&   1.242  & 0.219  & 0.9396  & 0.7695\\





Trans+VR$_{\text{RoBERTa}}$   & 0.792 &  0.644 & 0.504 & 0.386 & 0.280  &  0.582 & 1.244 & 0.219 & 0.9395 & 0.7705 \\ 
Trans+VR$_{\text{RoBERTa-Object}}$ & 0.792 &  0.644 & 0.503 & 0.386 & 0.281  &  0.582 & 1.242 & 0.219 & 0.9396 & 0.7701 \\


\hline 
Trans+VR$_{\text{BERT-Multi-class}}$   & \textbf{0.794} & 0.645 & 0.503 & 0.385 & 0.281    &   0.582  &  1.248 &   \textbf{0.220} & 0.9395 & \textbf{0.7717} \\
Trans+VR$_{\text{BERT-Multi-object}}$   & 0.792 &0.644 & 0.502 &  0.385 &  0.281    &   0.582  &   1.244 &  \textbf{0.220} & 0.9395 & 0.7693 \\
Trans+VR$_{\text{RoBERTa-Multi-class}}$  & 0.791  &0.643 & 0.503 & 0.385 &   0.280    &   0.581   & 1.244    &  0.219 &    0.9395   & 0.7710\\ 
Trans+VR$_{\text{RoBERTa-Multi-object}}$  & 0.791 & 0.643& 0.502& 0.385&  0.281    &   0.582   & 1.243   &  0.219 &   0.9395 & 0.7712 \\ 

\hline 

Trans+VR$_{\text{RBERTa-Class+Object}}$  & 0.793 & 0.645 & 0.504 & 0.387 & 0.281 &  0.582 &1.247  & 0.220 &   0.9397  & 0.7705  \\ 

Trans+VR$_{\text{BERT-Class+Object}}$  &0.793  & 0.645 & \textbf{0.505} &  \textbf{0.388} & \textbf{0.282} &  \textbf{0.583} & \textbf{1.251} & \textbf{0.220} &   \textbf{0.9399}  & 0.7695 \\


\hline 

\end{tabular}
}


\caption{\textbf{Positive Evidence: full result with all evaluation metrics}. Performance of compared baselines on the Karpathy test split with/without semantic re-ranking. For each base system, we report performance using a greedy search and the best beam search. Re-ranking is applied to the top-20 results of each system using BERT or RoBERTa for caption-context similarity. The visual contexts are extracted using ResNet152 and the Inception Resnet v2 based Faster R-CNN \textbf{object} detector. We also report results for Bert-based similarity without a hypothesis probability (rows marked \textit{only sim}).}

\label{se:Table_app}
\end{table*}
\begin{table*}[t!]
\small 
\center 

\begin{tabular}{cccl}
\hline
VC$_{1}$ & VC$_{2}$ & VC$_{3}$ & Caption  \\ \hline
cheeseburger & plate & hotdog &   a plate with a hamburger fries and tomatoes \\ 
bakery & dining table & web site &  a table having tea and a cake on it \\ 
gown & groom  & apron & its time to cut the cake at this couples wedding \\ \ 
racket & scoreboard & tennis ball &a crowd is watching a tennis game being played \\ 
laptop & screen&  desktop computer&a grey kitten laying on a windows laptop \\ 
washbasin & toilet& seat tub& a bathroom toilet sitting on a stand next to a tub and sink \\ \hline 
\end{tabular}
\caption{\textbf{Training Dataset}. The visual context (VC) is from a pre-trained visual classifier (\ie ResNet152) and the caption is from COCO-Caption dataset (human-annotated).}

\label{tb:dataset-coco}
\end{table*}

\begin{table*}[ht]
\small 
\centering

\vspace{0.1cm}
\resizebox{\textwidth}{!}{
\begin{tabular}{l c c c c c ccccc}

 \hline 
Model & B-1  & B-2  & B-3 &  B-4 &  M &  R  & C & S & BERTscore& SBERT-sts \\
 \hline



\multicolumn{11}{l}{VilBERT \cite{lu202012}} \\ 
Vil$_{\text{Greedy}}$     & 0.751  & 0.587  & 0.441 & 0.330 & 0.272 & 0.554 &  1.104 & 0.207 &0.9352 & 0.7550 \\ 

\vspace{0.1cm}
Vil$_{\text{BeamS}}$     & 0.752 &  0.592  & 0.456    &  0.351  &  0.274 &  0.557   & 1.115 & 0.205 & 0.9363 & 0.7550 \\

Vil+VR$_{\text{W-Object}}$  \cite{fang2015captions} & \textbf{0.756} &  \textbf{0.595}  & 0.456    &  0.348  &  0.274 &  \textbf{0.559} & 1.123 &0.206 &  0.9365 &  \textbf{0.7605} \\ 
Vil+VR$_{\text{Object}}$   \cite{wang2018object}   & \textbf{0.756} &  0.594  &0.455  &  0.348  &  0.274 &   \textbf{0.559}   & 1.120 & 0.206 & 0.9364 & 0.7570 \\

Vil+VR$_{\text{Control}}$  \cite{cornia2019show}  & 0.753 &  0.591  & 0.453    &  0.345  &  0.274 &  0.557  & 1.116 & 0.206 & 0.9361 & 0.7565 \\

\rowcolor{Gray} 
Vil+VR$_{\text{RoBERTa}}$  Table \ref{se:Table} (positive)    & 0.753 & 0.594  & \textbf{0.458}   &  \textbf{0.353} &  \textbf{0.276}    &  \textbf{0.559}   & 1.128  & 0.207 & \textbf{0.9366}  &  0.7562 \\



Vil+VR$_{\text{RoBERTa}}^{-low}$    & 0.748 &  0.588  & 0.453    &  0.349  &  0.275 &  0.557  & 1.116 & 0.206  & 0.9362  & 0.7531 \\ 


Vil+VR$_{\text{RoBERTa}}^{-high}$    & 0.748 &  0.588  & 0.453    &  0.349  &   0.275 &  0.557  & 1.116  & 0.206  & 0.9364 &   0.7546 \\

Vil+VR$_{\text{GloVe}}^{-pos}$    & 0.751 &  0.591  & 0.455    &  0.351  &  0.276 &   0.558  & 1.123 & 0.207  & 0.9364  & 0.7556 \\


\rowcolor{aliceblue}
Vil+VR$_{\text{RoBERTa+GloVe}}^{-joint}$    & 0.750 &  0.591  & 0.455    &   0.351  &  \textbf{0.276} &   \textbf{0.559}  & \textbf{1.126} & \textbf{0.208}  &  \textbf{0.9365}&  0.7548 \\

\hline 
\multicolumn{11}{l}{Transformer Caption Generator \cite{Marcella:20}} \\ 
Trans$_{\text{Greedy}}$     & 0.787   & 0.634 & 0.488 & 0.368 & 0.276 & 0.574 &   1.211 & 0.215 & 0.9376 & 0.7649 \\ 
\vspace{0.1cm}
Trans$_{\text{\text{BeamS}}}$      & 0.793 &  0.645  & 0.504 &   0.387  &0.281 &  0.582   &  1.247 & \textbf{0.220} & \textbf{0.9399} & 0.7707 \\ 

Trans+VR$_{\text{W-Object}}$  \cite{fang2015captions} &  0.786 &   0.638  & 0.497    &  0.378  &  0.277 &   0.579 & 1.228 & 0.216 &  0.9388 & 0.7709 \\ 
Trans+VR$_{\text{Object}}$   \cite{wang2018object}   &  0.790 &   0.642  &  0.501    &  0.383 &   0.280 &0.580 &  1.237   & 0.219 &  0.9391 & 0.7700  \\ 
Trans+VR$_{\text{Control}}$  \cite{cornia2019show}   & 0.791 &  0.644  & 0.505    &   \textbf{0.388}   &  0.281 &  \textbf{0.583}  & 1.248 & \textbf{0.220} & 0.9398 & \textbf{0.7716} \\ 
\rowcolor{Gray} 
\vspace{0.1cm}
Trans+VR$_{\text{BERT}}$ Table \ref{se:Table} (positive)  & 0.793  &  \textbf{0.646}   &  \textbf{0.505} &  \textbf{0.388}   &  \textbf{0.282}  & \textbf{0.583} &   \textbf{1.250}  &  \textbf{0.220} & \textbf{0.9399}  & 0.7711 \\  

Trans+VR$_{\text{BERT}}^{-low}$   & 0.791 &   0.643  & 0.504    &  0.387  &  0.280 &  0.582  & 1.242 & 0.218   & 0.9396 & 0.7682 \\
Trans+VR$_{\text{BERT}}^{-high}$   & 0.793 &  0.644  & 0.503    &  0.385  &  0.282 & 0.582  &  1.243 & 0.219  &  0.9397 & 0.7686 \\


\rowcolor{aliceblue}
Trans+VR$_{\text{GloVe}}^{-pos}$   & \textbf{0.794} &  \textbf{0.646}  & \textbf{0.506}    &  \textbf{0.388}   &  \textbf{0.282} &  \textbf{0.583} & \textbf{1.249} & \textbf{0.220}  & \textbf{0.9399} & 0.7702 \\

Trans+VR$_{\text{BERT+GloVe}}^{-joint}$   & 0.793 & 0.645  & 0.504    &  0.387  &  0.281 &  0.582  & 1.247 & \textbf{0.220}  & 0.9398 & 0.7704   \\

\hline 

\end{tabular}
}

\caption{\textbf{Negative Evidence: full result with all evaluation metrics}. Comparison results between positive Belief Revision (single concept VR) (gray color) and Negative Belief Revision (NBR) on the Karpathy test split. The NBR uses a \textit{high similarity} VR$^{-high}$ object related to the positive visual but not in the image, \textit{low similarity}  VR$^{-low}$ uses false positive from the visual classifier, and positive visual via static word level similarity VR$^{-pos}$. \textbf{Boldface} fonts reflect the improvement over the baseline.}
\label{se:Table_neg}
\end{table*}

\begin{table*}[t!]\centering

\small 
\resizebox{\textwidth}{!}{
\begin{tabular}{lcccccccccc}


\hline 
 Model &  B-1 & B-2 & B-3 & B-4 & M & R & C  & S & BERTscore & SBERT-sts \\
\hline 


\multicolumn{6}{l}{VilBERT-VR-GPT-2$_{\text{mean}}$ + ResNet}	\\ 
\rowcolor{Gray}
+ RoBERTa (Table \ref{se:Table} Best)      & \textbf{0.753} & \textbf{0.594}  & \textbf{0.458}   &  \textbf{0.353} &  0.276  & 0.559  &1.128 & 0.207 & 0.9366 &  0.7562 \\

+ LM-GPT-2$_{\text{product}}$  &    0.749    	&  0.590 &  0.455	&  0.351 & 0.276 &  0.558 & 1.124 & 0.208 & 0.9364 & 0.7486\\


+ DistilSBERT  &     0.751    	&  0.591 &  0.456	&   0.352 &  \textbf{0.277}  &  0.559 & 1.130  & 0.209 & 0.9365 & 0.7567 \\


+ SimCSE-BERT  \cite{gao2021simcse} &  0.752  &  0.593   &  0.457  &  0.352   & 0.276   &  0.559  &   1.130  &  \textbf{0.209} &   0.9365  & 0.7558 \\ 
+ SimCSE-RoBERTa    &   0.750  &  0.590   &  0.455  &  0.351   & 0.276   &  0.558  &   1.125  &  0.208 &   0.9365 & 0.7549 \\  


 + SimCSE-BERT-V$_{1}$ (unspervised)  &       0.750  &   0.591  &  0.455 &   0.351   & 0.276  &  0.558  &   1.128  &  0.207 &    0.9365 & 0.7560    \\
+ SimCSE-BERT-V$_{2}$ (unspervised)  &      0.752  &   0.593  &  0.457 &  \textbf{0.353}   & \textbf{0.277}  &   0.559  &   1.132  &  0.208 &    0.9365 & 0.7560    \\






+ CLIP-V \cite{radford2021learning} &    \textbf{0.753}   	&   \textbf{0.594}  &   \textbf{0.458} 	& \textbf{0.353} & 0.276 &\textbf{0.561} &  \textbf{1.131}& 0.208&  \textbf{0.9367} &  \textbf{0.7579}    \\

\hline 
\addlinespace[0.1cm]
\multicolumn{6}{l}{Trans - VR-GPT-2$_{\text{mean}}$ + ResNet}	\\ 
\rowcolor{Gray}
+ BERT (Table \ref{se:Table} Best)  & 0.793  &  \textbf{0.646}   &  \textbf{0.505} &  \textbf{0.388}   &  \textbf{0.282}  & \textbf{0.583} &   \textbf{1.250}  &  \textbf{0.220} & \textbf{0.9399}  & \textbf{0.7711} \\  

+ LM-GPT-2$_{\text{product}}$   &    0.787    	& 0.642 & 0.503	&  0.386 & 0.279  & 0.581& 1.236  &  0.219 &  0.9398 & 0.7683\\

\addlinespace[0.1cm]


+ DistilSBERT  & \textbf{0.794} & \textbf{0.646} &  \textbf{0.505} &  0.387&    \textbf{0.282} &   \textbf{0.583}	&  1.247 & \textbf{0.220} &  0.9396 & 0.7704 \\


+ SimCSE-BERT \cite{gao2021simcse} &  0.792  &   0.644   &  0.503  &   0.386   & 0.281   &  0.581  &    1.243  &  0.219 &   0.9394  & 0.7694 \\  

+ SimCSE-RoBERTa   &   \textbf{0.794}  &  0.645   &  0.504  &  0.387   & 0.281   &  0.582  &   1.244  &  0.219 &   0.9395  & 0.7698 \\

+ SimCSE-V$_{1}$ (unspervised) & 0.792 & 0.645 &  0.504 & 0.386 & 0.281 & 0.582 & 1.244 & 0.219 & 0.9397 & 0.7705  \\ 
+ SimCSE-V$_{2}$ (unspervised)  & 0.792 & 0.645 &  \textbf{0.505} & 0.387 & 0.281 & 0.582 & 1.247 & 0.219 & 0.9396 &  0.7703 \\


+ CLIP  \cite{radford2021learning}                       &     0.791             &    0.643    &    0.503      &   0.386     &  0.280     &    0.581      &  1.242     &    0.219     & 0.9395 & 0.7703\\

\hline 

\end{tabular}
}

\caption{\textbf{Full Ablation Study}. We experimented using different information from various baselines on the Karpathy test split. Also, with the unsupervised BERT similarity, we tried with the top-2 visual context from the classifier.}


\label{tb:comparsion}
\end{table*}

\begin{table*}[]
\caption{Examples show caption re-ranked by our \textbf{V}isual \textbf{R}e-ranker with different evaluation metrics including the semantic-similarity based metrics. Note that,  we only report  B-1 and B-2, from BLEU, to measure the word level changes before and after re-ranking.}



 \vspace{-0.2cm}
\resizebox{\textwidth}{!}{
\centering
\begin{tabular}{llccccccc}
\hline
 Model &  caption & B1 & B2 & M & R-L &  S &    BERTscore  &  SBERT-sts \\ 
 \hline    
BeamS   &     a woman holding a tennis racquet on a tennis court   &  \textbf{0.54}    &  \textbf{0.40} & 0.26  &  0.47 & 0 & 0.89 &   \textbf{0.73} \\ 
 \vspace{0.1cm}
    VR       &  a woman \textbf{standing} on a tennis court holding a racque       & 0.53    & 0.32 & \textbf{0.27} & \textbf{0.49}  & \textbf{0.30} &  \textbf{0.93} & 0.68   \\ 
    Refe    & \multicolumn{8}{l}{a woman in a short bisque skirt holding a tennis racque} \\ 
\hline



BeamS     &    a white train is at a train station         &  0.29 &  0.18 & 0.11   & 0.32   & 0.22  &0.90  & 0.57  \\ 
 \vspace{0.1cm}
VR       &    \textbf{a train on the tracks} at a train station    &  \textbf{0.49} &   \textbf{0.40} &  \textbf{0.23}    &  \textbf{0.52} & \textbf{0.53} &  \textbf{0.91} &  \textbf{0.78}  \\ 

Refe    & \multicolumn{8}{l}{a train that is sitting on the tracks under wires} \\ 
    \hline 
 
BeamS     &     \textcolor{red}{a pair of black scissors on a white wall}    \color{red}{\xmark}    &  0.31 &  0  & 0.12    &  0.37 & 0    &  0.87  & 0.23  \\ 
 \vspace{0.1cm}
VR       &     a flower in a vase \textcolor{red}{next to a pair of scissors}  \color{red}{\xmark}   &  \textbf{0.41}   &   \textbf{0.19} &  \textbf{0.15}     &   \textbf{0.43}  &   0.42   & \textbf{0.89}  & \textbf{0.27} \\ 
Refe    & \multicolumn{8}{l}{a dried black flower in a long tall black and white vase} \\ 

      \hline

BeamS     &       a man sitting on a bench       &  0.43 &  0.43 & 0.32    & 0.67  & 0.50  & 0.96 &   0.66 \\ 
 \vspace{0.1cm}
VR       &       a man sitting on a bench \textbf{talking on a cell phone}   &  \textbf{0.90}  & \textbf{0.85}    &  \textbf{0.95}     &  \textbf{0.90} & \textbf{0.90} & \textbf{0.98}   &  \textbf{0.99} \\

Refe    & \multicolumn{7}{l}{a man sitting on a bench talking on his cell phone} \\ 

    \hline
BeamS     &       a woman standing in an airport with luggage       &  0.42 &  0.35 & 0.25   & 0.51    &0.30 &  \textbf{0.92} &   0.67 \\ 
 \vspace{0.1cm}
VR       &        a woman standing in a \textcolor{red}{luggage carousel} at an airport \color{red}{\xmark}  &   \textcolor{red}{0.54}  &  \textcolor{red}{0.40}    &  0.25      & \textcolor{red}{0.56} &   \textcolor{red}{0.50} & 0.89  & \textcolor{red}{0.68} \\ 
Refe    & \multicolumn{8}{l}{a woman standing in front of a bench covered in luggage} \\ 

    \hline
BeamS     &       an airplane sitting on a runway behind a fence       &  \textbf{0.39} &  \textbf{0.29}& 0.22   & 0.41 & 0.40   &  0.92  &   0.69 \\ 
 \vspace{0.1cm}
VR       &       an airplane is parked behind a fence   &  0.37   & 0.28     &  \textbf{0.25}      &   \textbf{0.45} & \textbf{0.50} & \textbf{0.94}  & \textbf{0.85} \\ 
 
Refe    & \multicolumn{8}{l}{the airplane has landed behind a fence with barbed wire} \\ 

    \hline

BeamS     &       a plate with a sandwich on a table       &   0.55  &   0.26  & 0.19    & \textbf{0.46}  & 0.40   &  0.91  &   0.86 \\ 
 \vspace{0.1cm}
VR       &       \textbf{a white plate} with a sandwich on a table   &  \textbf{0.66}   & \textbf{0.40}     &   \textbf{0.24}       &   0.44 & \textbf{0.54} &  \textbf{0.92}   & \textbf{0.90}  \\ 
Refe    & \multicolumn{7}{l}{a small sandwich sitting on a white china plate} \\ 

  \hline 
BeamS     & \textcolor{red}{three} giraffes standing in a field under a tree       &   \textbf{0.26}   &   0     &   \textbf{0.14}    & \textbf{0.29}   & \textbf{0.26}  &  \textbf{0.91}  &   0.62 \\ 
 \vspace{0.1cm}
VR        & a group of giraffes standing in a field               &   0.17   &   0     &   0.10     &   0.20  & 0 &  0.90    & \textbf{0.64}  \\ 

Refe    & \multicolumn{8}{l}{two tall giraffe standing next to a green leaf filled tree} \\ 

 \hline 
   
BeamS     & \textcolor{red}{two} parking meters in front of a brick wall       &   \textbf{0.33}  &   0.20    &  0.15  &  0.22   & 0.25  &  0.47   &   0.87 \\ 
 \vspace{0.1cm}
VR        &  a row of parking meters in front of a building             &   0.30   &   \textbf{0.25} &   \textbf{0.17}     &   \textbf{0.31} & 0.25 &  \textbf{0.60}   & \textbf{0.88}  \\ 
 
Refe    & \multicolumn{7}{l}{different types and sizes of parking meters on display} \\ 

 \hline 
 
BeamS     &  a bathroom with a toilet and a mirror       & 0.33    &   0    &  0.10  &  \textbf{0.34}    & 0  & 0.89   &   0.69\\ 
 \vspace{0.1cm}
VR        &  a variety of items on display in a bathroom            &   \textbf{0.44}   &   0 &   \textbf{0.12}     &   0.33 & \textbf{0.14}  &  \textbf{0.90}   & \textbf{0.70}  \\ 

Refe    & \multicolumn{8}{l}{a view of a couple types of toilet items} \\ 

     \hline

BeamS     &   two bulls with horns standing next to each other          &   \textbf{0.44}    &   \textbf{0.23}    &  0.16  &   \textbf{0.47}   &  \textbf{0.66} &  \textbf{0.89}      &   0.76 \\
 \vspace{0.1cm}
VR        &   two long horn bulls standing next to each other            &   0.33    &  0   &   \textbf{0.18}  &   0.23    & 0.25  &  0.88      &  \textbf{0.81}  \\

Refe    & \multicolumn{7}{l}{closeup of two red-haired bulls with long horns} \\ 

    \hline

BeamS     &    a laptop computer sitting on top of a desk         &   0.44    &   0.21    &  0.20  &   0.29     & 0.18  & 0.91      &   0.69 \\ 
 \vspace{0.1cm}
VR        &   a desk with a laptop \textbf{and a computer monitor}           &   \textbf{0.53}    &   \textbf{0.51}  &   \textbf{0.31}  &   \textbf{0.58}  &  \textbf{0.46}   &  \textbf{0.95}      &  \textbf{0.77} \\

Refe    & \multicolumn{7}{l}{an office desk with a laptop and a phone on it} \\ 

    \hline



          %

BeamS     &     a busy highway with cars and a train         &   \textbf{0.33}   &   \textbf{0.20}   &  \textbf{0.12}  &   \textbf{0.34}    &  \textbf{0.18}  &     \textbf{0.90}      &   \textbf{0.43} \\ 
 VR        &   cars are driving on a highway \textcolor{red}{under a bridge} \color{red}{\xmark}       &   0.22    &   0  &   0.04  &   0.22    & 0  &  0.88      &   0.21 \\

Refe    & \multicolumn{8}{l}{a photo of a train heading down the tracks} \\ 

    \hline

BeamS     &   a baby sitting in front of a cake         &   0.44   &   0.23    &  0.18  &   \textbf{0.46} & \textbf{0.36}   &  0.90      &   \textbf{0.81} \\ 
 \vspace{0.1cm}
VR        &   a baby sitting in front of a \textbf{birthday cake}        &   0.44    &   0.23  &   0.18  &   0.44      & 0.33 &  0.90      &  \textcolor{red}{0.77} \\

Refe    & \multicolumn{8}{l}{a baby in high chair with bib and cake} \\ 

    \hline       


BeamS     &  a dog sitting on a bed with clothes         &  \textbf{0.58}   &   \textbf{0.44}    &  \textbf{0.29}  &   \textbf{0.65}   &  0.40  & \textbf{0.93}      &   0.57 \\ 
 \vspace{0.1cm}
VR        &    a dog sitting on a bed \textbf{next to} clothes         &   0.49    &  0.33  &   0.25 &   0.52      &0.40  &  0.91      &  \textbf{0.61} \\

Refe    & \multicolumn{7}{l}{a dog is sitting on an unmade bed with pillows} \\ 

    \hline

BeamS     &  a group of boats docked in the water        &   0.19   &   0    &  0.07  &   \textbf{0.21}    & 0.22 &  0.88      &   0.58 \\ 
 \vspace{0.1cm}
VR        &  a group of boats \textbf{are} docked in the water   &   0.19    &   0  &   0.07  &   0.20     & 0.22 &  0.88      &  \textbf{0.59} \\

Refe    & \multicolumn{8}{l}{looking out over a bay with many tourist boats moored} \\ 

    \hline

\end{tabular}
}
\label{tb:no-picture}
\end{table*}

\section{Human Evaluation}
We conducted a human study to investigate human preferences over the visual re-ranked caption. We randomly selected 26 test images and gave 12 reliable human subjects the option to choose between two captions: (1) Best-beam (BeamS) and (2) \textbf{V}isual \textbf{R}-ranker  as shown in Figure \ref{fig:HR}.%

Also, inspired by BERTscore and following \cite{song2021sentsim} that introduce Sentence Semantic  Semantic (SSS) for machine translation, we employ a sentence-to-sentence semantic similarity score to compare candidate captions with human references. We use pre-trained Sentence-RoBERTa$_{\text{LARGE}}$ tuned for general STS-B task. Consequently, the embedding will be more robust semantically than lexically for the STS tasks. Figure \ref{fig:human} shows that our results with SBERT-sts  agrees more with human judgment than the BERTscore. Figure \ref{tb-fig:images-example-1} and Figure \ref{tb-fig:images-example} show some examples of when humans agree/disagree with our re-ranker.


\section{Hyperparameters and Setting}

All training and the beam search are implemented with PyTorch 1.7.1  \cite{paszke2019pytorch}. VR based BERT$_{\text{base}}$ is fine-tuned on the training dataset using the original BERT implementation, Tensorflow version 1.15 with Cuda 8 \cite{abadi2016tensorflow} (hardware: GPU GTX 1070Ti and 32 RAM and 8-cores i7 CPU). The textual information dataset consists of  460k captions, 373k for training, and 87k for validation \ie visual, caption, label ([semantically related or not related]). We use a batch size of 16 for two epochs with a learning rate $2\mathrm{e}{-5}$, we kept the rest of hyperparameters  settings as the original implementation.





\section{Additional Examples}

\begin{figure*}[t!]

\centering
\resizebox{\textwidth}{!}{
\begin{tabular}{ccccccc}
Model & Caption &   BERTscore & SBERT-sts  &  Human\% & Visual   \\
\hline   



 \Vcentre{BeamS} & \Vcentre{a close up of a plate of food} &  \Vcentre{0.89 \DrawPercentageBarNeg{0.89}}   & \Vcentre{0.27 \DrawPercentageBarNeg{0.27}}  & \Vcentre{40 \DrawPercentageBarNeg{0.4}}   & \Vcentre{}  trifle \\ 
 \vspace{-0.4cm}
\Vcentre{VR} & \Vcentre{piece of food sitting on top of a white plate} &   \Vcentre{\textbf{0.91}  \DrawPercentageBar{0.91}}  & {\Vcentre{\textbf{0.53 \DrawPercentageBar{0.53}}}}   & \Vcentre{\textbf{60} \DrawPercentageBar{0.6}} &
   \Vcentre{\includegraphics[width = 2 cm, height = 2 cm]{COCO_val2014_000000467250.jpg}} \\ 
\vspace{0.07cm}
 \Vcentre{Human refe} & \Vcentre{a white plate and a piece of white cake} & \Vcentre{} \\

\hline 
    \Vcentre{BeamS} & \Vcentre{a group of men on a field playing baseball} &   \Vcentre{0.88 \DrawPercentageBarNeg{0.88}}&  \Vcentre{0.58 \DrawPercentageBarNeg{0.58}} & \Vcentre{33.3 \DrawPercentageBarNeg{0.33}} & \Vcentre{}  baseball \\ 
 \vspace{-0.4cm}
\Vcentre{VR} & \Vcentre{a batter catcher and umpire during a baseball game} &  \Vcentre{\textbf{0.91}  \DrawPercentageBar{0.91}} & \Vcentre{\textbf{0.84}  \DrawPercentageBar{0.84}}    & \Vcentre{\textbf{66.7}  \DrawPercentageBar{0.66}} &
   \Vcentre{\includegraphics[width = 2 cm, height = 2 cm]{COCO_val2014_000000202923.jpg}} \\ 
\vspace{0.07cm}
 \Vcentre{Human refe} & \Vcentre{batter catcher and umpire anticipating the next pitch} & \Vcentre{} \\ 
 
 \hline

  \Vcentre{BeamS} & \Vcentre{a couple of airplanes that are flying in the sky} &  \Vcentre{0.88  \DrawPercentageBarNeg{0.88}} &  \Vcentre{0.03 \DrawPercentageBarNeg{0.03}} &  \Vcentre{\textcolor{white}{00}0  \DrawPercentageBarNeg{0}} & \Vcentre{} traffic light  \\ 
 \vspace{-0.4cm}
\Vcentre{VR} & \Vcentre{ a group of traffic lights at an airport} &   \Vcentre{\textbf{0.95}  \DrawPercentageBar{0.95}} & \Vcentre{\textbf{0.71}  \DrawPercentageBar{0.71}}   & \Vcentre{\textbf{100}  \DrawPercentageBar{1}}   & 
   \Vcentre{\includegraphics[width = 2 cm, height = 2 cm]{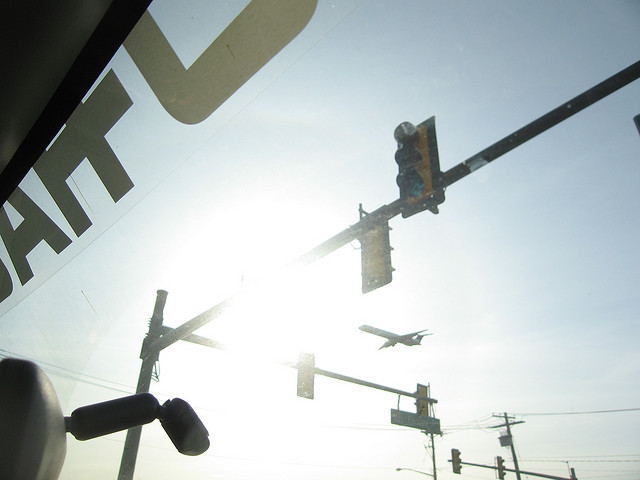}} \\ 
\vspace{0.07cm}
 \Vcentre{Human refe} & \Vcentre{a group of traffic lights sitting above an intersection} & \Vcentre{} \\ 
 
 \hline

   \Vcentre{BeamS} & \Vcentre{two bulls with horns standing next to each other} &   \Vcentre{\textbf{0.89}   \DrawPercentageBar{0.89}}  &  \Vcentre{0.76   \DrawPercentageBarNeg{0.76}} & \Vcentre{16.7   \DrawPercentageBarNeg{0.16}}  &  \Vcentre{}  ox \\ 
 \vspace{-0.4cm}
\Vcentre{VR} & \Vcentre{two long horn bulls standing next to each other} &  \Vcentre{0.88   \DrawPercentageBarNeg{0.88}} &   \Vcentre{\textbf{0.81}   \DrawPercentageBar{0.81}}  & \Vcentre{\textbf{83.3}   \DrawPercentageBar{0.83}} & 
   \Vcentre{\includegraphics[width = 2 cm, height = 2 cm]{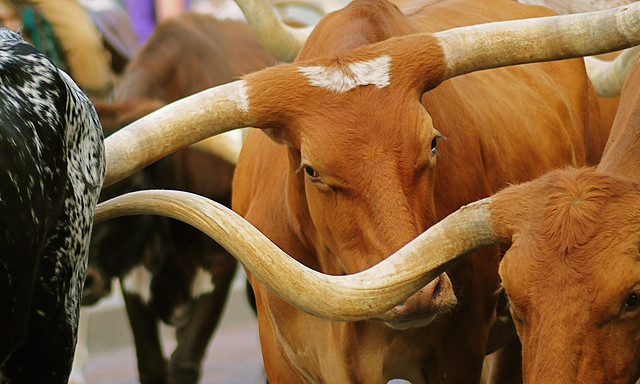}} \\ 
\vspace{0.07cm}
 \Vcentre{Human refe} & \Vcentre{closeup of two red-haired bulls with long horns} & \Vcentre{} \\ 
 \hline 
 
 

 
 

 \Vcentre{BeamS} & \Vcentre{a woman holding a tennis racquet on a tennis court} &  \Vcentre{0.89   \DrawPercentageBarNeg{0.89}} & \Vcentre{\textbf{0.73}   \DrawPercentageBar{0.73}} &  \Vcentre{\textbf{85.3}  \DrawPercentageBar{0.85}} & \Vcentre{}  racket \\ 
 \vspace{-0.4cm}
\Vcentre{VR} & \Vcentre{ a woman standing on a tennis court holding a racquet} &  \Vcentre{\textbf{0.93}   \DrawPercentageBar{0.93}} & \Vcentre{0.68   \DrawPercentageBarNeg{0.68}}  & \Vcentre{16.7  \DrawPercentageBarNeg{0.16}} & 
   \Vcentre{\includegraphics[width = 2 cm, height = 2 cm]{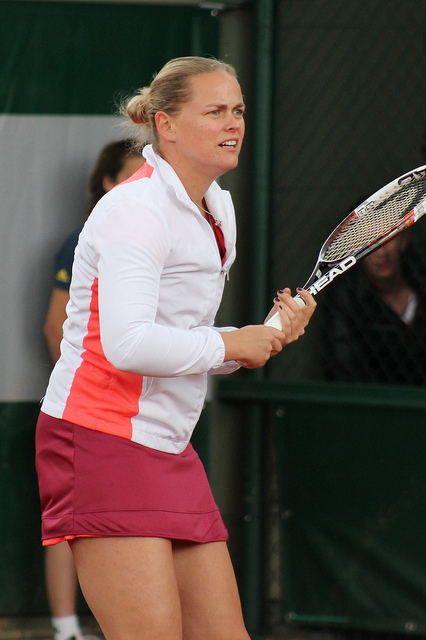}}  \\ 
\vspace{0.07cm}
 \Vcentre{Human refe} & \Vcentre{a woman in a short bisque skirt holding a tennis racquet} & \Vcentre{} \\

    \hline

\Vcentre{BeamS} & \Vcentre{a white train is at a train station} &  \Vcentre{0.90   \DrawPercentageBarNeg{0.90}}&   \Vcentre{0.57   \DrawPercentageBarNeg{0.57}}  & \Vcentre{50   \DrawPercentageBar{0.5}}   &   \Vcentre{} \text{\faBolt}  locomotive  \\ 
 \vspace{-0.4cm}
\Vcentre{VR} & \Vcentre{a train on the tracks at a train station} & \Vcentre{\textbf{0.91}   \DrawPercentageBar{0.91}} &  \Vcentre{\textbf{0.78}   \DrawPercentageBar{0.78}}      & \Vcentre{50   \DrawPercentageBar{0.50}}   &
   \Vcentre{\includegraphics[width = 2 cm, height = 2 cm]{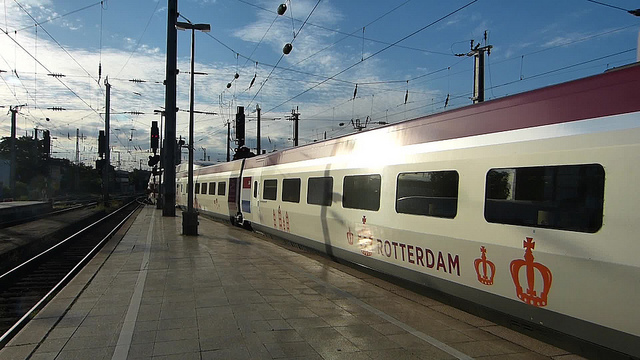}} \\ 
\vspace{0.07cm}
 \Vcentre{Human refe} & \Vcentre{a train that is sitting on the tracks under wires} & \Vcentre{} \\ 




 
 \hline  
 
  \Vcentre{BeamS} & \Vcentre{two men cutting a cake at ceremony} &  \Vcentre{\textbf{0.94}   \DrawPercentageBar{0.94}}  & \Vcentre{\textbf{0.98}   \DrawPercentageBar{0.98}} &   \Vcentre{\textbf{66.7}   \DrawPercentageBar{0.66}} & \Vcentre{}  $\approx$ mortarboard \\ 
 \vspace{-0.4cm}
\Vcentre{VR} & \Vcentre{a group of military men cutting a cake} &  \Vcentre{0.80 \DrawPercentageBarNeg{0.8}} &  \Vcentre{0.91 \DrawPercentageBarNeg{0.91}}  & \Vcentre{33.3 \DrawPercentageBarNeg{0.33}}  &  \Vcentre{\includegraphics[width = 2 cm, height = 2 cm]{COCO_val2014_000000048332.jpg}} \\ 

\vspace{0.07cm}
 \Vcentre{Human refe} & \Vcentre{two men are cutting a cake at a function} & \Vcentre{} \\

   \hline
 
\Vcentre{BeamS} & \Vcentre{a little girl wearing a tie and pants} &  \Vcentre{\textbf{0.94} \DrawPercentageBar{0.94}} &   \Vcentre{\textbf{0.80} \DrawPercentageBar{0.80}}  & \Vcentre{\textbf{66.7} \DrawPercentageBar{0.66}}   &  \Vcentre{ }  $\approx$ feather boa \\ 
 \vspace{-0.4cm}
\Vcentre{VR} & \Vcentre{a little girl wearing a tie standing in a room} & \Vcentre{0.93 \DrawPercentageBarNeg{0.93}} &  \Vcentre{0.77 \DrawPercentageBarNeg{0.77}}      & \Vcentre{33.3 \DrawPercentageBarNeg{0.33}}   &
   \Vcentre{\includegraphics[width = 2 cm, height = 2 cm]{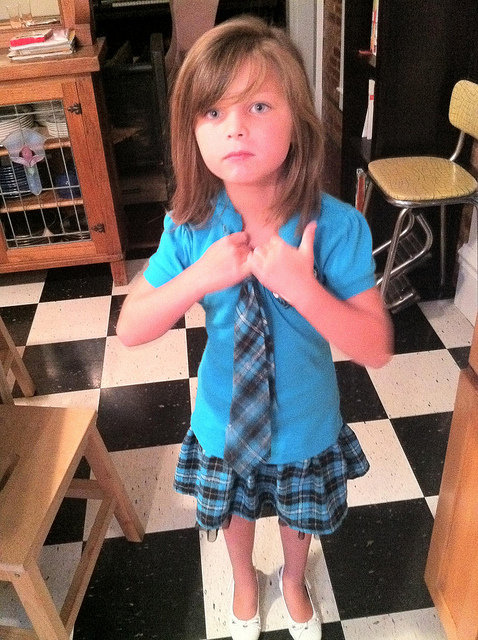}} \\ 
\vspace{0.07cm}
 \Vcentre{Human refe} & \Vcentre{a young girl wearing a tie that matches her skirt} & \Vcentre{} \\

   \hline

   \Vcentre{BeamS} & \Vcentre{a man laying on the ground with many animals} &  \Vcentre{0.88\DrawPercentageBarNeg{0.88}}  &  \Vcentre{0.14 \DrawPercentageBarNeg{0.14}} & \Vcentre{\textcolor{white}{00}0   \DrawPercentageBarNeg{0}} & \Vcentre{} \color{red}{\xmark} \textcolor{black}{trilobite}   \\ 
 \vspace{-0.4cm}
\Vcentre{VR} & \Vcentre{a man kneeling down in front of a herd of sheep} &   \Vcentre{\textbf{0.89} \DrawPercentageBar{0.89}} &  \Vcentre{\textbf{0.56} \DrawPercentageBar{0.56}}    & \Vcentre{\textbf{100} \DrawPercentageBar{1}}  & 
   \Vcentre{\includegraphics[width = 2 cm, height = 2 cm]{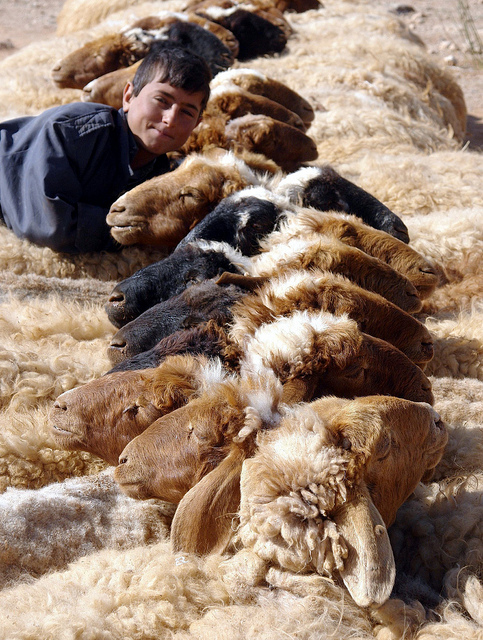}} \\ 
\vspace{0.07cm}
 \Vcentre{Human refe} & \Vcentre{a view of a bunch of sheep lined up with a behind them} & \Vcentre{} \\

   \hline

   \Vcentre{BeamS} & \Vcentre{a kitchen with black counter tops and wooden cabinets} &  \Vcentre{0.88  \DrawPercentageBar{0.88}} &  \Vcentre{\textbf{0.44}  \DrawPercentageBar{0.44}}  & \Vcentre{\textbf{100}  \DrawPercentageBar{1}} & \Vcentre{}  \color{red}{\xmark} \textcolor{black}{barbershop}  \\ 
 \vspace{-0.4cm}
\Vcentre{VR} & \Vcentre{ a kitchen counter with a black counter top} &   \Vcentre{0.88  \DrawPercentageBar{0.88}}   &\Vcentre{0.40  \DrawPercentageBarNeg{0.4}} &  \Vcentre{\textcolor{white}{00}0   \DrawPercentageBarNeg{0}} & 
   \Vcentre{\includegraphics[width = 2 cm, height = 2 cm]{COCO_val2014_000000293611.jpg}} \\ 
\vspace{0.07cm}
 \Vcentre{Human refe} & \Vcentre{a kitchen with a sink bottles jars and a dishwasher} & \Vcentre{} \\ 
 
\hline 

 


\end{tabular}
}
\caption{Examples show caption re-ranked by our \textbf{V}isual \textbf{R}e-ranker  and the original baseline Best beam. An evaluation metrics comparison between semantic-similarity based SBERT-sts, BERTscore, and the human subject.}
\label{tb-fig:images-example-1}

\end{figure*}

\begin{figure*}[t!]

\centering
\resizebox{\textwidth}{!}{
\begin{tabular}{ccccccc}
Model & Caption &   BERTscore & SBERT-sts  &  Human\% & Visual   \\

\hline

 \Vcentre{BeamS} & \Vcentre{a red and white boat in the water} &    \Vcentre{\textbf{0.94}  \DrawPercentageBar{0.94}}   &   \Vcentre{\textbf{0.78}  \DrawPercentageBar{0.78}}  &   \Vcentre{\textbf{87.5}  \DrawPercentageBar{0.87}} & \Vcentre{}  fireboat \\ 
 \vspace{-0.4cm}
\Vcentre{VR} & \Vcentre{a red and white boat is in the water} &   \Vcentre{0.93  \DrawPercentageBarNeg{0.93}}  & \Vcentre{0.79  \DrawPercentageBarNeg{0.79}}   & \Vcentre{12.5  \DrawPercentageBarNeg{0.12}} &
   \Vcentre{\includegraphics[width = 2 cm, height = 2 cm]{COCO_val2014_000000330369.jpg}} \\ 
\vspace{0.07cm}
 \Vcentre{Human refe} & \Vcentre{a red and white boat floating along a river} & \Vcentre{} \\

\hline 
    \Vcentre{BeamS} & \Vcentre{a dog sitting on a bed with cloths} &     \Vcentre{\textbf{0.93}  \DrawPercentageBar{0.93}} &  \Vcentre{0.57  \DrawPercentageBarNeg{0.57}} & \Vcentre{50  \DrawPercentageBar{0.5}} & \Vcentre{}  dog/Irish terrier  \\ 
 \vspace{-0.4cm}
\Vcentre{VR} & \Vcentre{a dog sitting on a bed next to clothes} &   \Vcentre{0.91  \DrawPercentageBarNeg{0.91}} & \Vcentre{\textbf{0.61}  \DrawPercentageBar{0.61}}    &   \Vcentre{50  \DrawPercentageBar{0.5}} &
   \Vcentre{\includegraphics[width = 2 cm, height = 2 cm]{COCO_val2014_000000022192.jpg}} \\ 
\vspace{0.07cm}
 \Vcentre{Human refe} & \Vcentre{a dog is sitting on an unmade bed with pillows} & \Vcentre{} \\ 
 
 \hline

  \Vcentre{BeamS} & \Vcentre{a laptop computer sitting on top of a desk} &  \Vcentre{0.91  \DrawPercentageBarNeg{0.91}} &  \Vcentre{0.69  \DrawPercentageBarNeg{0.69}} &  \Vcentre{25  \DrawPercentageBarNeg{0.25}} &  desk  \\ 
 \vspace{-0.4cm}
\Vcentre{VR} & \Vcentre{a desk with a laptop and computer monitor}  &    \Vcentre{\textbf{0.95  \DrawPercentageBar{0.95}}} &   \Vcentre{\textbf{0.77}  \DrawPercentageBar{0.77}}   &   \Vcentre{\textbf{75}  \DrawPercentageBar{0.75}}   & 
   \Vcentre{\includegraphics[width = 2 cm, height = 2 cm]{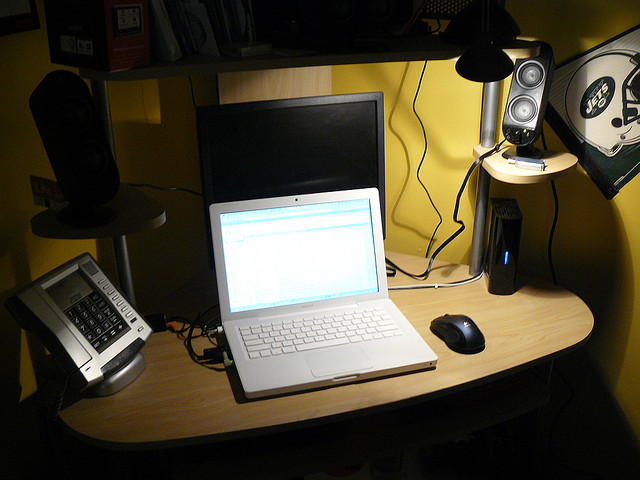}} \\ 
\vspace{0.07cm}
 \Vcentre{Human refe} & \Vcentre{an office desk with a laptop and computer monitor} & \Vcentre{} \\ 
 
 \hline

   \Vcentre{BeamS} & \Vcentre{a bathroom with a toilet and a mirror} &    \Vcentre{0.89  \DrawPercentageBarNeg{0.89}}  &  \Vcentre{0.69  \DrawPercentageBarNeg{0.69}} & \Vcentre{12.5  \DrawPercentageBarNeg{0.12}}  &  \Vcentre{}  washbasin \\ 
 \vspace{-0.4cm}
\Vcentre{VR} & \Vcentre{a variety of item on display in a bathroom} &  \Vcentre{\textbf{0.90}  \DrawPercentageBar{0.90}} &    \Vcentre{\textbf{0.70}  \DrawPercentageBar{0.70}}  &  \Vcentre{\textbf{83.3}  \DrawPercentageBar{0.83}} & 
   \Vcentre{\includegraphics[width = 2 cm, height = 2 cm]{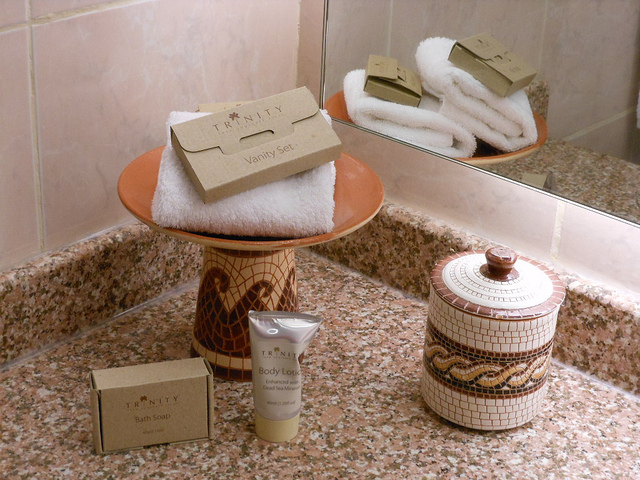}} \\ 
\vspace{0.07cm}
 \Vcentre{Human refe} & \Vcentre{a view of a couple types of toilet items} & \Vcentre{} \\ 
 \hline 
 
 
\Vcentre{BeamS} & \Vcentre{a couple of pizzas that are on a table} & \Vcentre{\textbf{0.88}  \DrawPercentageBar{0.88}}  & \Vcentre{0.75 \DrawPercentageBarNeg{0.75}}  &  \Vcentre{\textbf{100}  \DrawPercentageBar{1}} &  \Vcentre{}  pizza \\ 

\vspace{-0.4cm}
\Vcentre{VR} & \Vcentre{ a couple of pizzas are sitting on a table} &  \Vcentre{0.87 \DrawPercentageBarNeg{0.87}} &  \Vcentre{\textbf{0.77}  \DrawPercentageBar{0.77}}   & \Vcentre{\textcolor{white}{00}0   \DrawPercentageBar{0}} & 
   \Vcentre{\includegraphics[width = 2 cm, height = 2 cm]{COCO_val2014_000000039671.jpg}} \\ 
\vspace{0.07cm}
 \Vcentre{Human refe} & \Vcentre{pizza on a table with a cup and a fork} & \Vcentre{} \\ 
 \hline 
 
 
 
\Vcentre{BeamS} & \Vcentre{ a group of people in a living room playing a video game} &  \Vcentre{0.93  \DrawPercentageBarNeg{0.93}}&   \Vcentre{0.47  \DrawPercentageBarNeg{0.47}}  & \Vcentre{\textbf{62.5}  \DrawPercentageBar{0.62}}   &   \Vcentre{} television  \\ 
 \vspace{-0.4cm}
\Vcentre{VR} & \Vcentre{a group of people sitting in a living room playing a video game} & \Vcentre{\textbf{0.94}  \DrawPercentageBar{0.94}} &  \Vcentre{\textbf{0.50}  \DrawPercentageBar{0.50}}      & \Vcentre{37.5  \DrawPercentageBarNeg{0.37}}   &
   \Vcentre{\includegraphics[width = 2 cm, height = 2 cm]{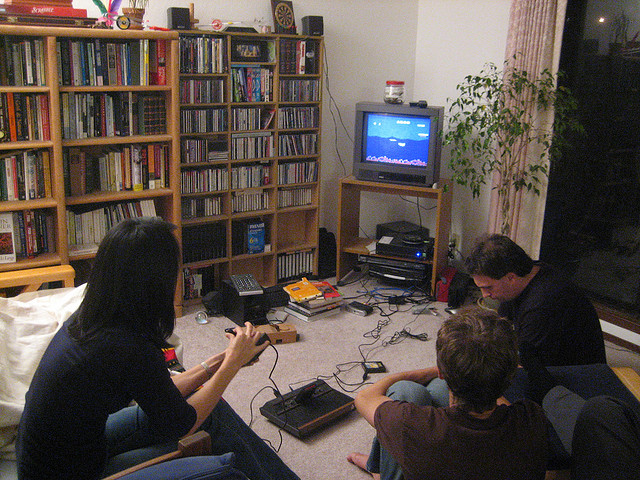}} \\ 
\vspace{0.07cm}
 \Vcentre{Human refe} & \Vcentre{a group of friends sitting inside their living room} & \Vcentre{} \\ 

 \hline

\Vcentre{BeamS} & \Vcentre{a city street at night with traffic lights} &  \Vcentre{\textbf{0.97  \DrawPercentageBar{0.97}}} &   \Vcentre{0.81 \DrawPercentageBarNeg{0.81}}  & \Vcentre{33.3 \DrawPercentageBarNeg{0.33}}   &  \Vcentre{} traffic light  \\ 
\vspace{-0.4cm}
\Vcentre{VR} & \Vcentre{a city street at night with cars and a traffic light} & \Vcentre{0.94 \DrawPercentageBarNeg{0.94}} &  \Vcentre{\textbf{0.85}  \DrawPercentageBar{0.85}}      & \Vcentre{\textbf{66.7}  \DrawPercentageBar{0.66}}   &
\Vcentre{\includegraphics[width = 2 cm, height = 2 cm]{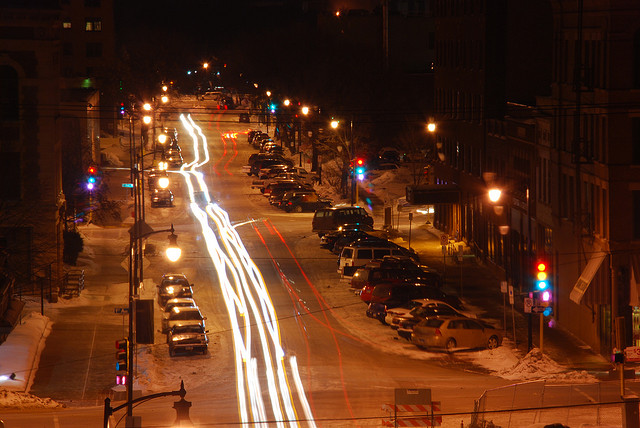}} \\ 
\vspace{0.07cm}
\Vcentre{Human refe} & \Vcentre{a city street at night filled with lots of traffic} & \Vcentre{} \\

 \hline 
 

    \Vcentre{BeamS} & \Vcentre{a close up of a cat eating a doughnut} & \Vcentre{\textbf{\textbf{0.83}}  \DrawPercentageBar{0.83}} & \Vcentre{\textbf{0.90}  \DrawPercentageBar{0.90}} & \Vcentre{\textbf{100}  \DrawPercentageBar{1}}  &  \Vcentre{} pretzel   \\ 
 \vspace{-0.4cm}
\Vcentre{VR} & \Vcentre{ a close up of a person holding a doughnut} &  \Vcentre{0.60  \DrawPercentageBarNeg{0.60}} & \Vcentre{0.88  \DrawPercentageBarNeg{0.88}}   & \Vcentre{\textcolor{white}{00}0   \DrawPercentageBarNeg{0}}   &
   \Vcentre{\includegraphics[width = 2 cm, height = 2 cm]{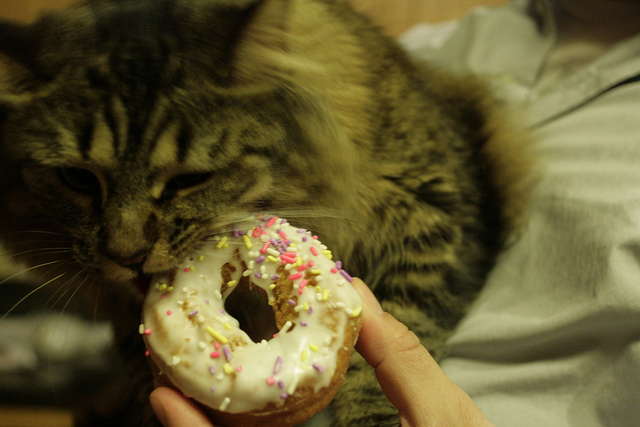}} \\ 
\vspace{0.07cm}
 \Vcentre{Human refe} & \Vcentre{a cat bites into a doughnut offered by a persons hand} & \Vcentre{} \\ 
 
 




   \hline

\Vcentre{BeamS} & \Vcentre{ two men standing next to a group of people} &  \Vcentre{\textbf{0.88}  \DrawPercentageBar{0.88}}  &  \Vcentre{0.20  \DrawPercentageBarNeg{0.20}} & \Vcentre{\textbf{87.5}  \DrawPercentageBar{0.87}} & \Vcentre{  \textcolor{black}{}} groom  \\ 
 \vspace{-0.4cm}
\Vcentre{VR} & \Vcentre{ a group of men standing next to each other} &   \Vcentre{0.86  \DrawPercentageBarNeg{0.86}} & \Vcentre{\textbf{0.30  \DrawPercentageBar{0.30}}}    & \Vcentre{12.5  \DrawPercentageBarNeg{0.12}}  & 
   \Vcentre{\includegraphics[width = 2 cm, height = 2 cm]{COCO_val2014_000000089487.jpg}} \\ 
\vspace{0.07cm}
\Vcentre{Human refe} & \Vcentre{the man is holding his tie with his right hand} & \Vcentre{} \\ 

\hline

 

 
  \Vcentre{BeamS} & \Vcentre{a pile of trash sitting inside of a building} &   \Vcentre{0.88  \DrawPercentageBar{0.88}} & \Vcentre{\textbf{0.38} \DrawPercentageBar{0.38}}  & \Vcentre{\textbf{100}  \DrawPercentageBar{1}}  & \Vcentre{ \textcolor{black}{}} \color{red}{\xmark} \textcolor{black}{vacuum}  \\ 
 \vspace{-0.4cm}
\Vcentre{VR} & \Vcentre{a pile of trash sitting in front of a building} & \Vcentre{0.88  \DrawPercentageBar{0.88}}  &  \Vcentre{0.27  \DrawPercentageBarNeg{0.27}}   & \Vcentre{\textcolor{white}{00}0   \DrawPercentageBarNeg{0}} & 
   \Vcentre{\includegraphics[width = 2 cm, height = 2 cm]{COCO_val2014_000000386613.jpg}} \\ 
\vspace{0.07cm}
 \Vcentre{Human refe} & \Vcentre{an older floor light sits deserted in an abandoned hospital} &  \\

 \hline

\end{tabular}
}



\caption{Examples show caption re-ranked by our \textbf{V}isual \textbf{R}e-ranker  and the original baseline Best beam. An evaluation metrics comparison between semantic-similarity based SBERT-sts, BERTscore, and the human subject.}

\label{tb-fig:images-example}

\end{figure*}

We provide more comparison results with examples, including sentence-level evaluation SBERT-sts and BERTscore  in Table  \ref{tb:no-picture}. Also, in  Figure  \ref{tb-fig:images-example-1}, and Figure \ref{tb-fig:images-example}, we also evaluated  our re-ranker using human subjects.

\begin{tabular}{@{} l l l @{}}
\end{tabular}

\end{document}